\documentclass[11pt,oneside]{article}

\usepackage[left=1in,right=1in,top=1in,bottom=1in]{geometry}
\usepackage{authblk}

\usepackage{setspace}
\usepackage[parfill]{parskip}
\usepackage{lscape}

\usepackage{color}
\usepackage[colorlinks,
            citecolor=blue,
            urlcolor=blue,
            linkcolor=blue]{hyperref}

\usepackage[utf8]{inputenc} 
\usepackage[T1]{fontenc}    
\usepackage{hyperref}       
\usepackage{url}            
\usepackage{booktabs}       
\usepackage{amsfonts}       
\usepackage{nicefrac}       
\usepackage{microtype}      
\usepackage{enumitem}
\usepackage{multirow}
\usepackage{algorithm}
\usepackage{nameref}
\usepackage{comment}

\usepackage{microtype}
\usepackage{graphicx}
\usepackage{subfig}
\usepackage{amsmath}   
\usepackage{amsthm}
\usepackage{amssymb}
\usepackage{siunitx}   
\usepackage{bm}        
\usepackage{thmtools}  
\usepackage{thm-restate}
\usepackage{color}     
\usepackage{xspace}    

\newcommand{\bw}{\bm{w}}
\newcommand{\bx}{\bm{x}}
\newcommand{\by}{\bm{y}}

\newcommand{\bX}{\bm{X}}

\newcommand{\Ccal}{\mathcal{C}}
\newcommand{\Fcal}{\mathcal{F}}
\newcommand{\Hcal}{\mathcal{H}}
\newcommand{\Gcal}{\mathcal{G}}

\newcommand{\R}{\mathbb{R}}
\newcommand{\Q}{\mathbb{Q}}
\newcommand{\E}{\mathbb{E}}

\newcommand{\bone}{\bm{1}}

\DeclareMathOperator*{\argmin}{arg\,min\,}
\DeclareMathOperator*{\argmax}{arg\,max\,}
\DeclareMathOperator{\Accuracy}{\textsf{\small Accuracy}}
\DeclareMathOperator{\Fairness}{\textsf{\small Fairness}}
\DeclareMathOperator{\Bias}{\textsf{\small Bias}}
\DeclareMathOperator{\conv}{\textsf{\small conv}}
\DeclareMathOperator{\Cov}{Cov}

\definecolor{ForestGreen}{RGB}{34,139,34}

\newcommand{\fairstacks}{\textsc{FairStacks}\xspace}

\newcommand{\taf}{\textsc{taf}\xspace}
\newcommand{\auc}{\textsc{auc}\xspace}
\newcommand{\fauc}{\textsc{fauc}\xspace}
\newcommand{\tafi}{\textsc{tafi}\xspace}
\newcommand{\fauci}{\textsc{fauci}\xspace}

\newtheorem{theorem}{Theorem}

\newtheorem{definition}{Definition}

\newtheorem{remark}{Remark}

\usepackage[natbib=true,maxcitenames=2,maxbibnames=20]{biblatex}
\renewbibmacro{in:}{}

\title{To the Fairness Frontier and Beyond:\\ Identifying, Quantifying, and Optimizing the Fairness-Accuracy Pareto Frontier}

\author[1]{Camille Olivia Little\footnote{These authors contributed equally.}}
\author[2]{Michael Weylandt$^*$}
\author[1,3,4]{Genevera I. Allen}
\affil[1]{Department of Electrical and Computer Engineering, Rice University}
\affil[2]{University of Florida Informatics Institute}
\affil[3]{Department of Computer Science, Rice University}
\affil[3]{Department of Statistics, Rice University}

\begin{document}

\maketitle

\begin{abstract}
Algorithmic fairness has emerged as an important consideration when developing and deploying machine learning models to make high-stakes societal decisions. Yet, improved fairness often comes at the expense of model accuracy. While aspects of the fairness-accuracy tradeoff have been studied, most work reports the fairness and accuracy of various models separately; this makes model comparisons nearly impossible without a unified model-agnostic metric that reflects the Pareto optimal balance of the two desiderata. In this paper, we seek to identify, quantify, and optimize the empirical Pareto frontier of the fairness-accuracy tradeoff, defined as the highest attained accuracy at every level of fairness for a collection of fitted models. Specifically, we identify and outline the empirical Pareto frontier through our Tradeoff-between-Fairness-and-Accuracy (\taf) Curves; we then develop a single metric to quantify this Pareto frontier through the weighted area under the \taf Curve which we term the Fairness-Area-Under-the-Curve (\fauc). Our \taf Curves provide the first empirical, model-agnostic characterization of the Pareto frontier, while our \fauc provides the first unified metric to impartially compare model families in terms of both fairness and accuracy. Both \taf Curves and \fauc are general and can be employed with all group fairness definitions and accuracy measures. Next, we ask: Is it possible to expand the empirical Pareto frontier and thus improve the \fauc for a given collection of fitted models? We answer in the affirmative by developing a novel fair model stacking framework, \fairstacks. \fairstacks solves a convex program to maximize the accuracy of a linear combination of fitted models subject to a constraint on score-based model bias. We show that optimizing with \fairstacks always expands the empirical Pareto frontier and improves the \fauc; we additionally study other theoretical properties of our proposed approach. Finally, we empirically validate \taf, \fauc, and \fairstacks through studies on several real benchmark data sets, showing that \fairstacks leads to major improvements in \fauc that outperform existing algorithmic fairness approaches.


\end{abstract}

\clearpage
\onehalfspace

\begin{refsection}

\section{Introduction} \label{sec:introduction}

Machine learning algorithms are now widely used to help make high-stakes decisions, such as deciding if an applicant should be approved for a loan or predicting if a convict will commit another crime. These decisions can have life-altering consequences and many have shown that machine learning models can be biased and unintentionally discriminate against protected groups \citep{Dastin:2018, Angwin:2016}.  In response to this major problem, many authors have developed techniques to identify and mitigate  bias in machine learning algorithms \citep{chouldechova:2018, Dwork:2012,Kamiran:2009, Berk:2017,Feldman:2015, Calmon:2017, Kamiran:2012, Kamiran:post}.  Many of these techniques can dramatically improve fairness, but often at the expense of model accuracy; this leads to what has been called the \emph{fairness-accuracy tradeoff} \citep{Zhao:2019,Zhao:2022, Zliobaite:2015, Menon:2018, Agarwal:2018, Martinez:2019}.  

When making certain high-stakes decisions, balancing the tradeoff between fairness and accuracy is absolutely critical. Take, for example, the Federal Fair House Act (FFHA) of 1968. The FFHA protects people from discrimination when engaging in housing-related activities such as renting or buying a home or getting a mortgage. The FFHA is designed to prohibit discrimination on the basis of race, religion, sex, and other protected attributes, yet as of 2020, Black and Latino mortgage applicants are more likely to be declined than white applicants \citep{Quillian:2018}. When considering loan applicants, lenders must balance FFHA requirements (fairness) and credit risk (accuracy). This example demonstrates a special case of more general fairness-accuracy tradeoff often seen in real-world scenarios.

In this work, our objective is to identify, measure and optimize the empirical Pareto frontier of the fairness-accuracy tradeoff. This Pareto frontier is defined as the highest attained accuracy at every level of fairness for a collection of fitted models. We are particularly motivated to define a model-agnostic measure that quantifies the fairness-accuracy tradeoff in a single metric in order for machine learning practitioners to easily compare existing bias mitigation strategies and choose hyperparameters that balance the tradeoff in a way suits their specific task.  Moreover, we seek to develop a flexible meta-learner that expands the empirical Pareto frontier for the set of models by optimizing the accuracy attainable at every level of fairness.  Again, our goal is to develop a model-agnostic approach that will improve fairness for any collection of models.  Taken together, these objectives would provide users simple and practical tools to assess and measure their models' Pareto frontier as well as an approach to further expand their models' frontier and improve fairness.

\subsection{Related Work}\label{subsec:related_work}
The existence of a fairness-accuracy tradeoff has been noted and partially characterized in several previous papers \citep{Menon:2018, Zhao:2019, Zhao:2022,Chen:2018, Zliobaite:2015}. \citeauthor{Zhao:2022} \citep{Zhao:2019,Zhao:2022} prove that, when base rates differ across protected groups, the minimum possible error of any fair classifies is bounded below by the difference in base rates. This gives a valuable theoretical bound on the extremal fairness-accuracy tradeoff, but does not give a concrete proposal for comparing different classifiers nor guidance on how to tune particular classifiers.

On the other hand, \citet{Menon:2018} consider the problem of finding an optimal decision rule, after a base classifier has been learned. They show that under a specific loss function, the resulting fairness-accuracy Pareto-frontier can be theoretically characterized, but their finding is not applicable to general classifiers. Additionally, these previous characterizations of the fairness-accuracy tradeoff are restricted to binary classification \citep{Zhao:2019,Zhao:2022, Menon:2018,Chen:2018, Zliobaite:2015}. 

A related line of work \citep{Agarwal:2018, Martinez:2020} attempts to explicitly optimize the fairness-accuracy Pareto frontier in different settings. \citet{Agarwal:2018} propose a model-agnostic method that iteratively calls a black-box model and reweights or relabels the data to find the most accurate fair version of the input classifier. The user can input any base learner into this method and recover the entire fairness-accuracy Pareto frontier using their constrained optimization model that minimizes error subject to a fairness constraint. This method offers a useful way to construct the frontier, but it takes as input only one class of models at a time. As a result, the quality of the frontier is limited by the highest achievable accuracy or fairness of the single input base learner. \citet{Martinez:2020} formulate group fairness as a multi-objective optimization problem where each group risk is an objective function. Though this provides a useful way to identify the Pareto classifier that minimizes risk of the worst performing group, it is not apparent how their method would extend to other group fairness notions.

Other proposals \citep{Kamishima:2012, Zafar:2017, Wu:2019, Martinez:2019, Berk:2017} attempt to find the fairest classifier on a certain problem by modifying existing classifiers to reduce discrimination, typically by some sort of fairness regularizer or constraint. For example, \citet{Kamishima:2012} introduce a regularization term to penalize discrimination when formulating a logistic regression classifier. Existing fairness-regularized methods work well in certain scenarios, but they suffer from three main limitations: i) the added fairness constraint typically yields a non-convex objective, posing significant computational challenges; ii) the approaches are \emph{ad hoc} and only applicable to specific model families; and iii) the practical guarantees associated with these relaxations are often insufficient in practice. \citep{Lohaus:2020}.

Existing bias mitigation strategies can be generally categorized into pre-processing techniques \citep{Kamiran:2012,Feldman:2015, Calmon:2017, Choi:2020, Jiang:2020}, in-processing techniques \citep{Zhang:2018, Berk:2017, Grari:2019, Celis:2019}, and post-processing techniques \citep{Hardt:2016, Pleiss:2017,Kamiran:post, Lohia:2019, Chzhen:2019}. Though they work independently from the data and the model, post-processing techniques often lead to a drastic decrease in accuracy as the results of the trained model are directly altered. To avoid this, we leverage popular approaches from ensemble learning by using model stacking \citep{Wolpert:1992, Dzeroski:2004} to find the fairness-accuracy Pareto frontier. Ensemble learning improves performance by reducing variance of the prediction error by adding more bias \citep{Dietterich:2000}: we will show that fairness constraints can similarly improve performance.
To our knowledge, no existing ensemble learning or other post-processing strategy stacks a set of models to specifically optimize the empirical fairness-accuracy Pareto frontier. 
 
 \begin{landscape}
\begin{figure}[htb]
\centering
\includegraphics[width=0.5\textwidth]{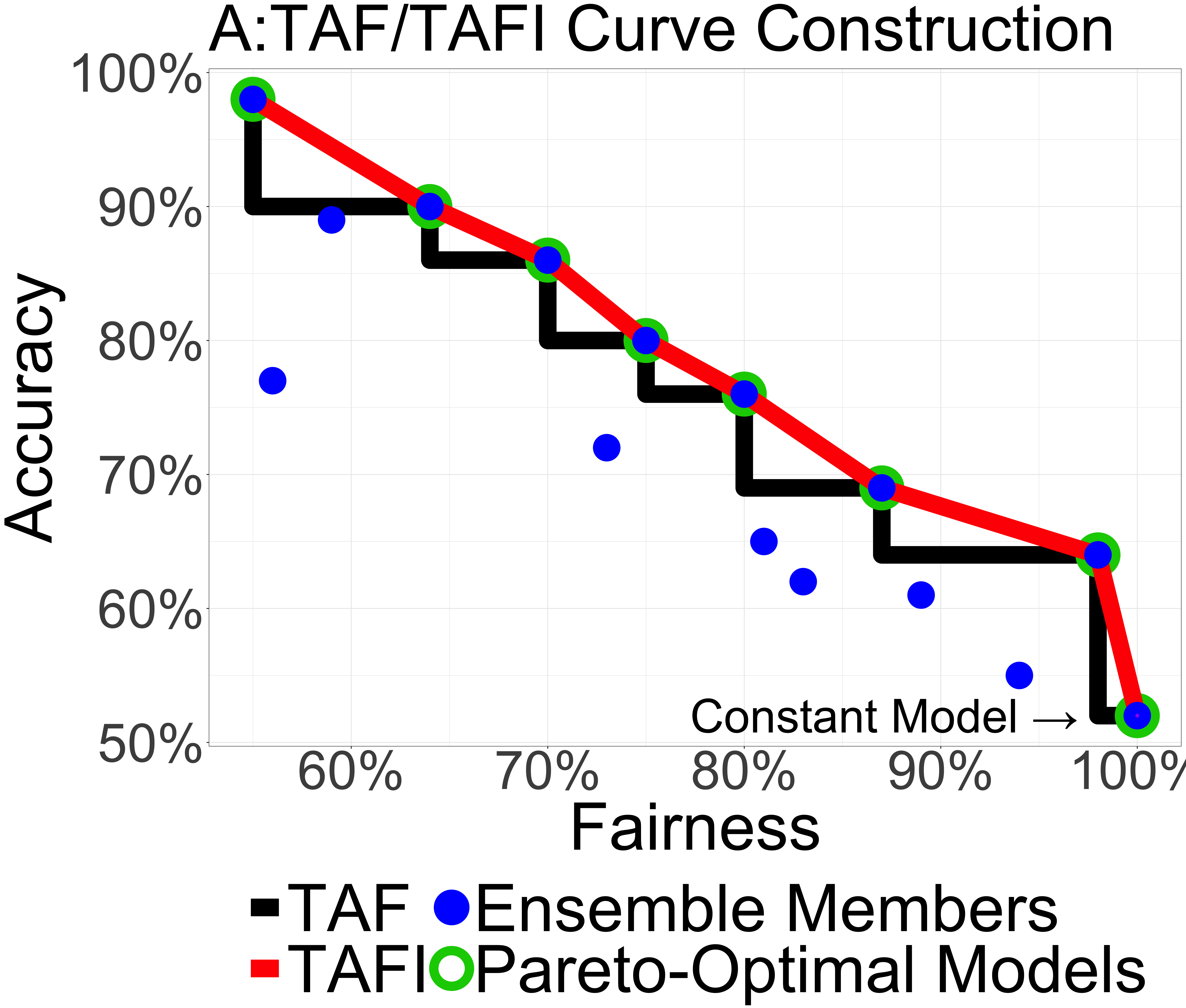} ~ \includegraphics[width=0.5\textwidth]{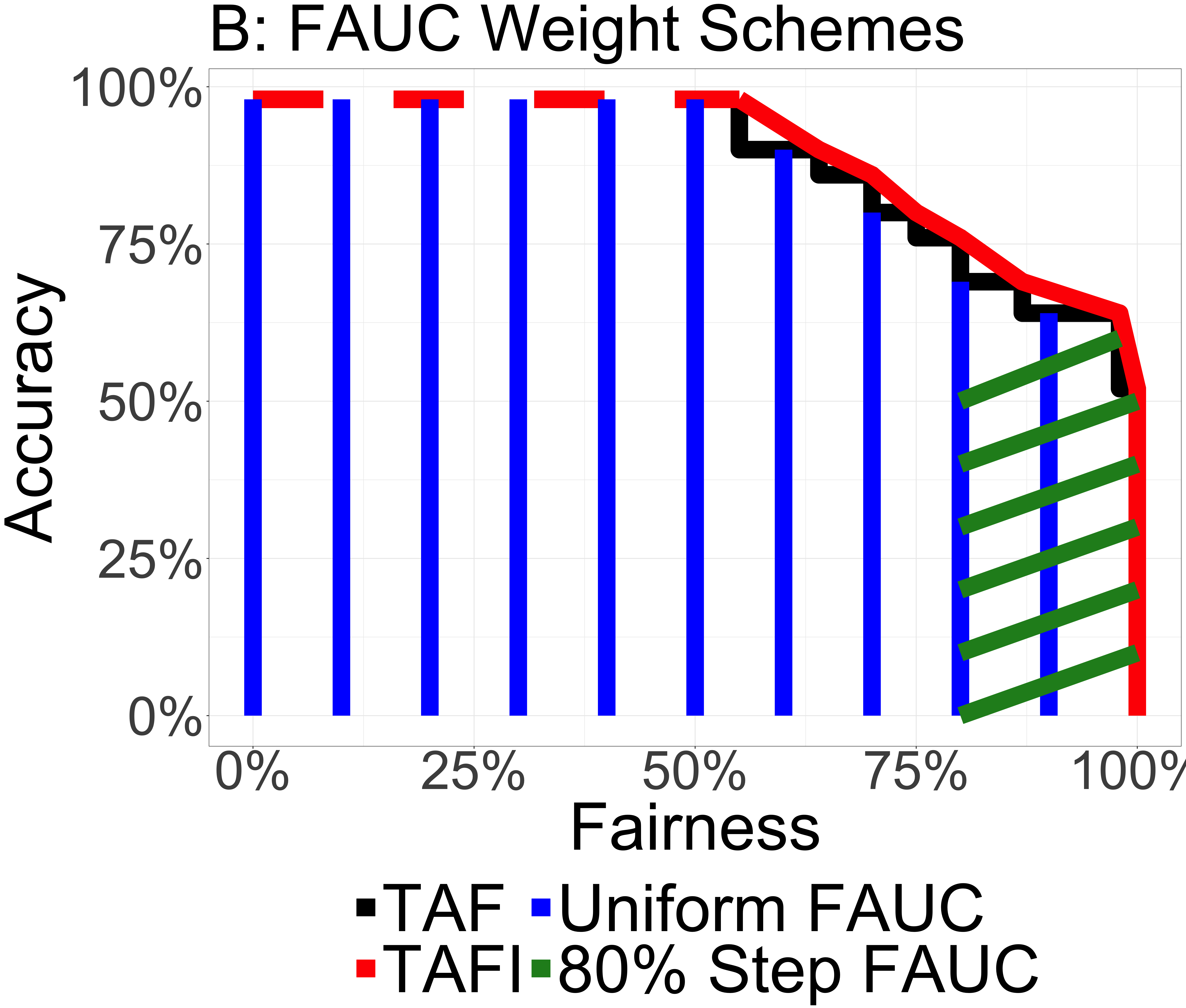} \\ \vspace{0.2in}~ \\
\includegraphics[width=0.5\textwidth]{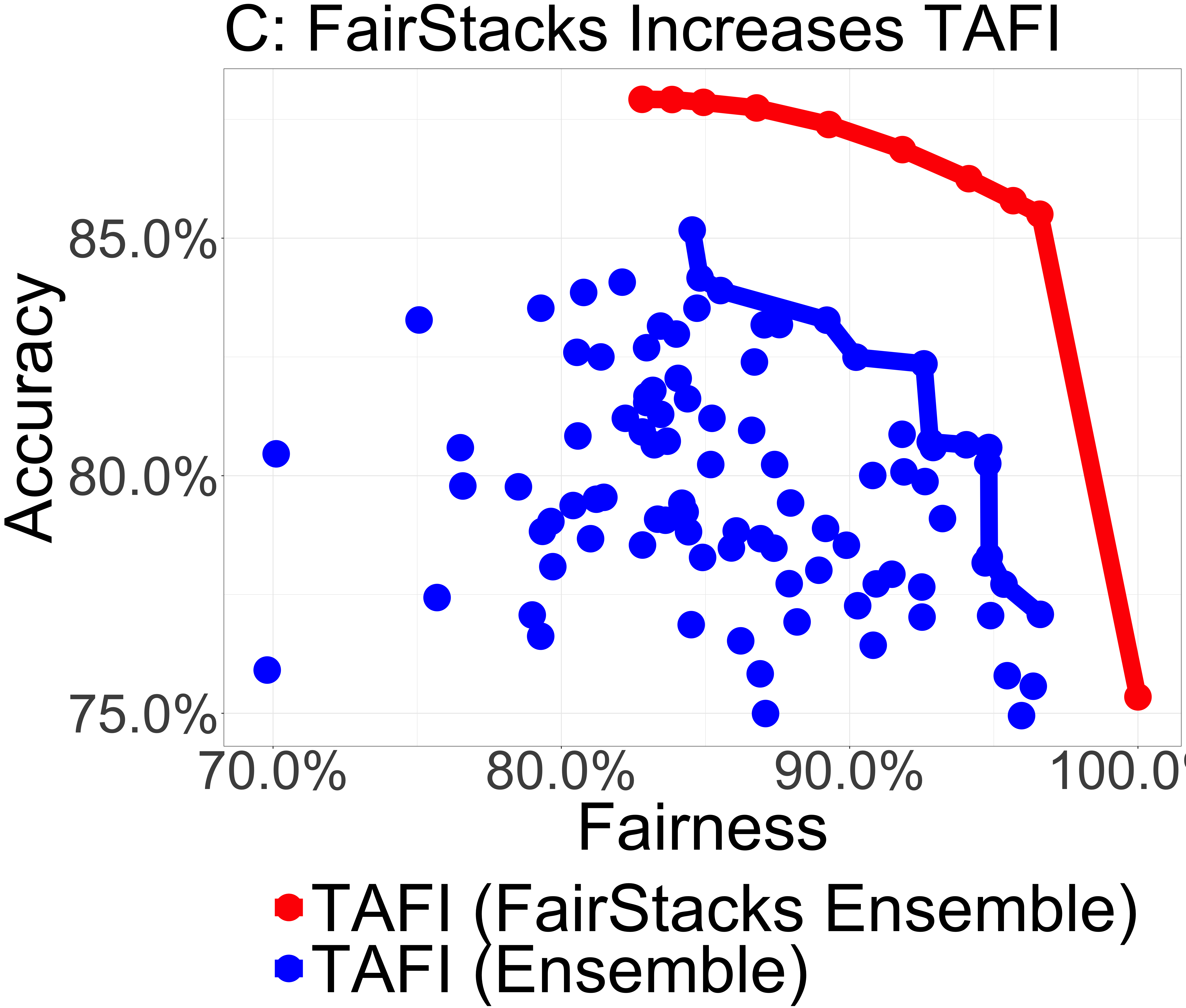} ~ \includegraphics[width=0.5\textwidth]{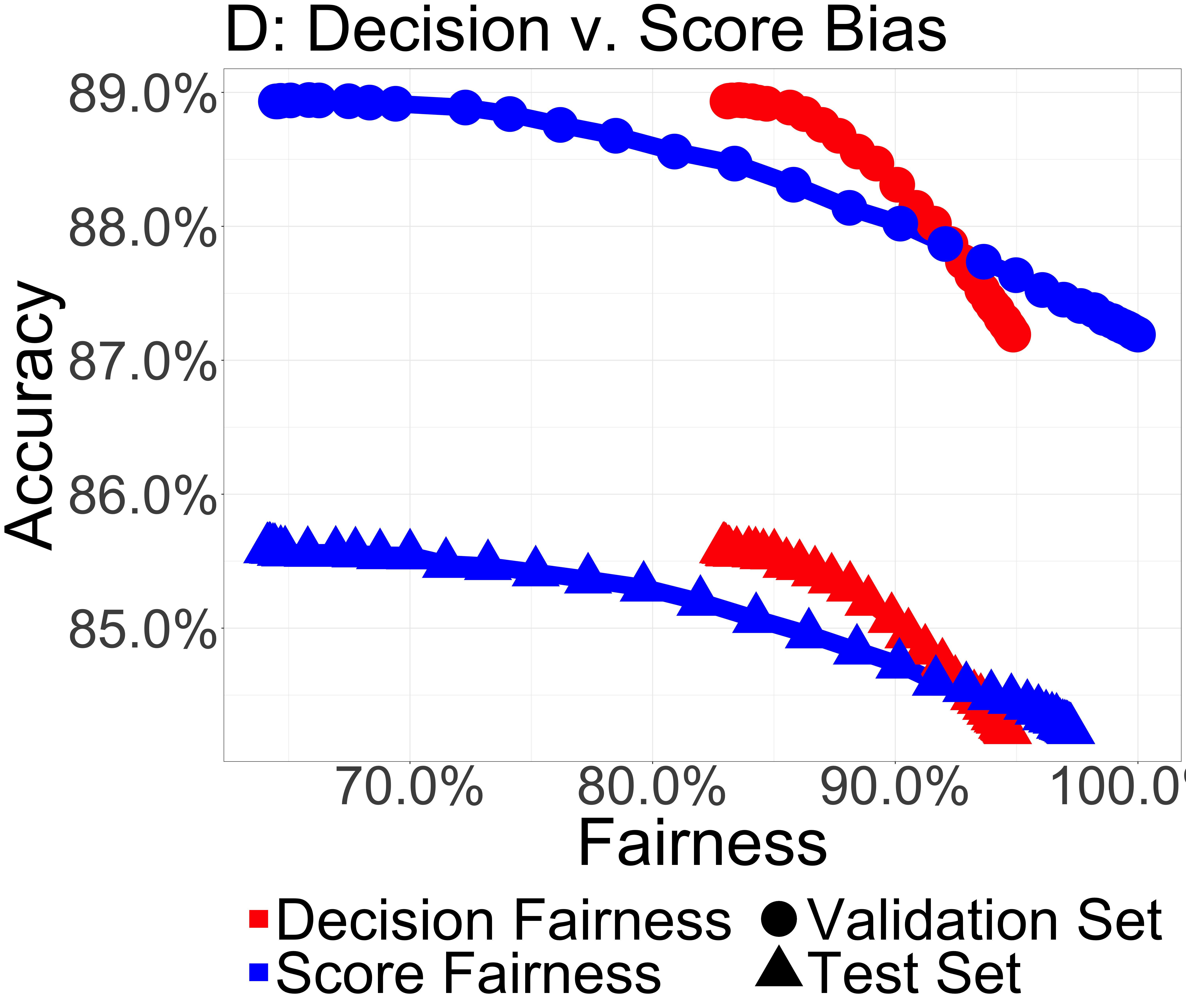} \\ 
\caption{Depictions of our major theoretical and methodological contributions. Panel A highlights the relationship between the model class $\Hcal$, its Pareto optimal members, and the resulting \taf and \tafi curves (Definitions \ref{def:paretomodel}-\ref{def:taf}); Panel B visualizes the \fauc metric for both the uniform and 80\% step-weight used in Section \ref{sec:empirical}; Panel C depicts how our \fairstacks procedure (Section \ref{sec:optimizing}) significantly improves upon the \taf of the individual ensemble members (Proposition \ref{prop:fairstackshelps}); and Panel D depicts the relationship between our proposed score bias and more standard measures of bias.} 
\label{fig:schematics}
\end{figure}
\end{landscape}
\subsection{Our Contributions}\label{subsec:contributions}

\begin{enumerate}
    \item We introduce the {\it Tradeoff-between-Fairness-and-Accuracy (\taf) Curve} which outlines the empirical Pareto frontier consisting of the highest attained accuracy within a collection of fitted models at every level of fairness.  This provides the {\it first model-agnostic quantitative and visual summary of the fairness-accuracy tradeoff} for a model family or other collection of models.  
    \item Next, we provide the {\it first unified, scalar and model-agnostic measure of the empirical fairness-accuracy tradeoff} by computing the weighted area under the \taf Curve, termed the {\it Fairness Area-Under-the-Curve} (\fauc).  Just as \auc is used to compare classifiers, \fauc is the first quantitative metric to impartially compare model families in terms of both accuracy and fairness. 
    \item Finally, we seek to expand the empirical Pareto frontier of the collection of fitted models through a novel {\it convex, bias-constrained model stacking framework called \fairstacks}. We show that under certain assumptions, {\it \fairstacks expands the frontier} and improves the \fauc, leading to a simple strategy to improve both the fairness and accuracy for any collection of models.    
    \item We empirically validate our metrics and \fairstacks framework on several real benchmark data sets, showing that with sufficient input models, \fairstacks dominates all approaches, providing the highest accuracy at all levels of fairness (\emph{i.e.}, dominating the fairness-accuracy Pareto frontier).  
\end{enumerate}

\section{Identifying the Frontier: Tradeoff-between-Fairness-Accuracy Curve} \label{sec:identifying}

We begin our study of the fairness-accuracy tradeoff by recalling the basic notions of Pareto improvement, Pareto optimality, and of the Pareto frontier. Specifically, given two options, $\Ccal_1$ and $\Ccal_2$, we say that $\Ccal_1$ is a \emph{Pareto improvement} over $\Ccal_2$ if $\Ccal_1$ is preferred over $\Ccal_2$ by any consumer. An option $\Ccal_1$ is said to be \emph{Pareto optional} if there exists no attainable Pareto improvement over it. The set of all Pareto optimal options is known as the \emph{Pareto frontier} \citep{MasColell:1995}. In this paper, we evaluate models on two metrics, fairness and accuracy, and the notion of a Pareto optimal model is easily characterized: 
\begin{definition} \label{def:paretomodel}
Given a set of models $\Hcal$, an accuracy metric $\Accuracy: \Hcal \to [0, 1]$, and a fairness metric $\Fairness: \Hcal \to [0,1]$ both of which can be written as sums (averages) over individual observations, we say that $h \in \Hcal$ is \emph{Pareto optimal in $\Hcal$} if there does not exist $h' \in \Hcal$ such that $\Accuracy(h') \geq \Accuracy(h)$ and $\Fairness(h') \geq \Fairness(h)$ with one inequality strict.
\end{definition}
Here, $\Hcal$ is any collection of models, $\Fairness(h)$ is any metric in $[0,1]$ that is decreasing in the bias of model $h$ according to standard fairness definitions \citep{Verma:2018}, and $\Accuracy(h)$ is any measure of the performance of model $h$ in $[0,1]$.  The collection of models, $\Hcal$ can be created by varying the hyperparameters of a fixed model family, by an ensemble of models, or by considering a larger collection of possible models for a given problem.  In what follows, we also assume that $\Hcal$ always contains a perfectly fair model, \emph{i.e.}, a model such that $\Fairness(h) = 1$. This assumption is trivially satisfied by including the constant (intercept-only) model and exists only to simplify the statements of our results.  In the context of classification, most definitions of accuracy take values in $[0,1]$ and can be used for $\Accuracy(\cdot)$, while for regression tasks, one may rescale typical losses to be in $[0,1]$ (e.g. $R^2$ or $e^{-\text{MSE}}$).  Most standard definitions of bias in algorithmic fairness take values in $[0,1]$ and hence we can let $\Fairness(h) = 1 - \Bias(h)$; as an example, we can define the fairness for Demographic Parity \citep{Dwork:2012,Feldman:2015} as $\Fairness_{\text{DP}}(h) = 1 - |\E(h | A = 1) - \E(h | A=0)|$, where $A$ denotes the protected attribute.  Overall, the machinery developed in this paper is very general and works for any definitions of fairness and accuracy taking values in $[0,1]$ that follow the notion of more is better; that is, more accuracy is preferred and more fairness is preferred.  

For a given collection of models $\Hcal$, it clearly does not make sense to use a model $h$ that is not Pareto optimal.  Thus, we seek to identify all Pareto optimal models in $\Hcal$.  Note that the collection of all of these optimal models forms the Pareto frontier; if these models are ordered, they also form an explicit fairness-accuracy tradeoff curve \emph{i.e.}, the empirical Pareto frontier.  This motivates the following: \begin{definition} \label{def:taf} Given a finite collection of models $\Hcal$, the \taf curve associated with $\Hcal$ is a function from $f \in [0,1]$ ($\Fairness$) to $[0,1]$ ($\Accuracy$) such that 
\[\taf_{\Hcal}(f) = \max_{h \in \Hcal_f} \Accuracy(f) \text{ where } \Hcal_f = \{ h \in \Hcal: \Fairness(h) \geq f\} \subseteq \Hcal.\]
\end{definition}
Informally, the \taf curve is the curve obtained by constructing a (left) step-function interpolation of the Pareto optimal members of $\Hcal$ and represents the best possible accuracy that can be obtained at a given level of fairness. 
Algorithm \ref{alg:one} details how, given a set of candidate models, the set of Pareto optimal models can be identified and the \taf curve constructed in time quasi-linear in the size of $\Hcal$.
\begin{remark} \label{remark:taf} Algorithm 1 identifies the set of all Pareto optimal models, $\Hcal^* \subseteq \Hcal$, and critical points of the \taf curve outlining the fair Pareto frontier of $\Hcal$.
\end{remark}
This remark follows directly from the definitions of Pareto optimality and the \taf Curve; further properties of our $\taf$ Curve are provided in the Supplement.  

As an illustration, consider Panel A and B in Figure~\ref{fig:schematics} (we plot accuracy and fairness in terms of percentages instead of in $[0,1]$).  The $\taf$ Curve for a collection of fitted models is a step function where each critical point is a Pareto optimal model.  The left end-point of the $\taf$ Curve is given by the highest accuracy obtained by any model in $\Hcal$ extended to $\Fairness(\cdot) = 0$; note that this left end-point is perhaps not practically relevant but follows from the definition of $\taf$.  The right end-point of the $\taf$ Curve is given by the constant model with fairness equal to one.  Hence, our $\taf$ Curve provides a way to determine the Pareto optimal models and Pareto frontier of any arbitrary collection of fitted models; this also provides a visual summary of the fairness-accuracy tradeoff for particular model classes as well as particular data sets.  

\begin{algorithm}[htb]
\caption{Compute Pareto Optimal Models and \taf Curve Points}\label{alg:one}
\vspace{0.05in} \textbf{Input:} Set of candidate models $\Hcal$, including $H$ arbitrary models and one perfectly fair model \\
\textbf{Output:} Ordered set of Pareto Optimal models $\Hcal^* \subseteq \Hcal$ and the step function points for \taf
\vspace{0.05in}
\hrule
\vspace{0.1in}
\begin{enumerate}
    \item \textbf{Pre-Compute:} Stably sort $\Hcal$ in descending order of fairness:
    \[\Fairness(h_{(0)}) \geq \Fairness(h_{(1)}) \geq  \Fairness(h_{(2)}) \geq \dots \geq \Fairness(h_{(H)})\]
    \item \textbf{Initialize:} $\Hcal^* = \{h_{(0)}\}$ where $h_{(0)}$ is the perfectly fair model
    \item \textbf{Filter:} For $i = 1, \dots, H$: 
    \begin{itemize}
        \item If $\Accuracy(h_{(i)}) > \max_{h \in \Hcal^*} \Accuracy(h)$: \begin{itemize}
            \item If $\Fairness(h_{(i)}) = \Fairness(h^*_{(|\Hcal^*|)})$, set $\Hcal^* = \Hcal^* \setminus \{h^*_{(|\Hcal^*|)}\}$
            \item Set $\Hcal^* = \Hcal^* \cup \{h_{(i)}\}$
        \end{itemize}
    \end{itemize}
    \item \textbf{Return:} 
    \begin{itemize}
        \item Pareto Optimal Models $\Hcal^*$
        \item \taf Curve Points: $\{(f, a): f = \Fairness(h^*), a = \Accuracy(h^*) \text{ for all } h^* \in \Hcal^*\}$
    \end{itemize}
\end{enumerate}
\end{algorithm}

\section{Quantifying the Frontier: Fairness Area-Under-the-Curve (\fauc)} \label{sec:quantifying}

Currently, many compare model families by reporting the fairness and accuracy for a single hyperparameter.  But given that there is often a tradeoff between fairness and accuracy, these metrics alone are not sufficient to determine the superior model or model family.  We thus turn to our \taf Curves and ask, can these be used to compare model families or other collections of models?  And, can we summarize our curves in a single, unified metric to facilitate comparisons?  The idea of using $\taf$ Curves, or the empirical Pareto frontier, naturally arises in many contexts: comparing model families on the same data set, comparing the level of bias for a fixed set of models in two data sets, or comparing the fairness of a fixed set of models before and after some intervention. Ideally, these comparisons are trivial, with one \taf Curve dominating the other at all fairness levels, but in practice \taf Curves often cross. To address these difficulties, we draw inspiration from ROC curves used to balance precision and recall in classification.  Here, people commonly use the Area-Under-the-ROC-Curve (\auc) to compare two or more ROC curves; the \auc gives a single metric summarizing the balance of both precision and recall.  Inspired by this, we propose to compute a simple scalar summary of the \taf Curve which we call the Fairness-Area-Under-the-Curve (\fauc):
\begin{definition}\label{def:fauc}
Given a finite collection of models $\Hcal$, an accuracy metric $\Accuracy: \Hcal \to [0, 1]$, a fairness metric $\Fairness: \Hcal \to [0,1]$, and a non-negative (measurable) weight function $w$, the \emph{$w$-weighted \fauc score of $\Hcal$} is $\fauc^w(\Hcal) = \left(\int_{[0, 1]} \taf_{\Hcal}(f) w(f) \,\mathrm{d}f\right) / \left(\int_{[0, 1]} w(f) \,\mathrm{d}f\right) \in [0,1]$. 
\end{definition}
Just like \auc, \fauc is a score between zero and one with one indicating perfect accuracy at all levels of fairness.  In practice, note that because the empirical \taf curve is piecewise constant, computing the \fauc score is a trivial (right-handed) Riemannian sum that can be computed in linear time.  

An illustration of our \fauc metric is given in Figure~\ref{fig:schematics}B.  Notice that since the \taf Curve definitionally extends the highest attained accuracy to left, this level of accuracy could dominate the \fauc score.  Since these associated lower levels of fairness often do not correspond to any model in the collection $\Hcal$, one may prefer to use a weighted $\fauc$, with non-zero weights for only the levels of fairness of interest or as needed for the particular context.  This motivates us to consider how the flexible weight function could be used to capture particular fairness preferences: 
\begin{restatable}{theorem}{faucutility} \label{prop:faucutility}
Suppose there exists a preference relation, $\succeq$, on the set of $\taf$ curves which is total, transitive, and continuous. Suppose further that the preference relation $\succeq$ is increasing in $\taf$, that is, if $\taf_{\Gcal} \geq \taf_{\Hcal}$ pointwise then $\taf_{\Gcal} \succeq \taf_{\Hcal}$, that the utility of accuracy is preferentially independent at any finite collection of fairness levels, and that the utility of constant \taf curves is linear in the accuracy. Then, there exists a non-negative generalized function $w$ such that $\taf_{\Gcal} \succeq \taf_{\Hcal} \Leftrightarrow \fauc^w_{\Gcal} \geq \fauc^w_{\Hcal}$ for all model sets $\Gcal, \Hcal$. 
\end{restatable}
This result utilizes economic preference theory (see \citep{MasColell:1995,Debreu:1983}) and uses standard assumptions in this field; the proof and a detailed discussion of the assumptions may be found in the Supplement. 

The implications of Theorem \ref{prop:faucutility} are profound: no matter how an individual would choose to trade-off accuracy and fairness, there is a \fauc variant that completely characterizes their views. For example, the preferences of an individual who cares only about accuracy and not at all about fairness are encoded by a point $w = \delta_0$ where $\delta_0$ is Dirac's delta and $\fauc^{\delta_0}(\Hcal) = \taf_{\Hcal}(0) = \text{max}_{h \in \Hcal} \Accuracy(h)$. We expect that, for most individuals, a weight function of the form $w(x) = x^{\alpha} \bone_{x > \beta}$ for some $\alpha \geq 1, \beta \in [0, 1]$ captures their preferences well, placing a premium on the accuracy of models that are (nearly-perfectly) fair and forcing adherence to some regulatory lower-bound on fairness, but we leave the question of precise utility elicitation to future work.  Overall, \fauc and its weighted variants provide an intuitive and  practical unified metric that summarizes the whole fairness-accuracy tradeoff.

\subsection{Improving the \fauc Score via Randomized Interpolation: \tafi and \fauci}\label{subsec:interpolation}

Randomizing aspects of model behavior has recently been shown to be an effective approach to improve fairness in ML systems \citep{Zhao:2020} and the \taf/\fauc framework can also benefit from randomization. For any two models $g, h$, we interpret the composite model $\alpha g + (1-\alpha) h$ for fixed $\alpha \in [0, 1]$ as the randomized procedure which applies $g$ with probability $\alpha$ and $h$ otherwise. When $g, h$ are Pareto optimal in $\Hcal$, this procedure produces models which are not Pareto dominated by any element of $\Hcal$. If these randomized combinations are added to $\Hcal$, the resulting \taf curve will consist of the Pareto optimal elements of $\Hcal$ linearly interpolated as shown in Figure \ref{fig:schematics}B: for this reason we refer to the resulting curve as the \taf + Interpolation curve, or more simply, \tafi, and the area under it as \fauci. These composite models can be considered as the \emph{convex hull} of $\Hcal$ and there is a connection between this convex hull of randomized procedures and the convex geometry of the \taf curve: 
\begin{restatable}{theorem}{convexfauc} \label{thm:convexfauc}
For a fixed collection of models $\Hcal$, 
$\tafi_{\Hcal} = \overline{\conv}(\taf_{\Hcal}) = \taf_{\conv(\Hcal)}$.  Hence, we have $\taf_{\Hcal} \leq \tafi_{\Hcal}$ pointwise and, by extension, $\fauc^w_{\Hcal} \leq \fauci^w_{\Hcal}$ for any weight function $w$. 
\end{restatable} 
Here, $\conv(\taf_{\Hcal})$ refers to the (geometric) convex hull of the \taf Curve, $\overline{A}$ denotes the upper boundary of a set $A$,  and $\taf_{\conv(\Hcal)}$ refers to \taf applied to the set of randomized procedures. While randomization is guaranteed to improve performance as measured by \fauc, we do note that randomization is sometimes at odds with transparency and auditability requirements, so this strategy is not universally applicable. Taking inspiration from Theorem \ref{thm:convexfauc} and the power of model combinations to improve \fauc, we next ask if we can optimize linear combinations of models to expand the fair Pareto frontier.

\section{Optimizing the Frontier: Fair Model Stacking (\fairstacks)} \label{sec:optimizing}

We have developed the \taf+\fauc framework for identifying and quantifying the empirical Pareto frontier attained by a collection of fitted models, but we ask: Is it possible to expand the frontier for these same set of models, perhaps using some meta-learner?  For this, we consider a meta-learner which uses linear combinations of models in a stacked ensemble learning framework.  Taking inspiration from common multi-objective optimization used to learn the Pareto frontier as well as recent work of this nature in the context of fairness \citep{Martinez:2020}, we propose to study the following problem:  
\begin{equation}
    \argmax_{\bw \in \R^k} \quad \Accuracy\left(\sum_{i = 1}^k w_i h_i(\bx); \by\right) 
    \text{ such that } \left|\Bias\left(\sum_{i=1}^k w_i h_i(\bx); \by\right)\right| \leq \tau
\label{eqn:general_stacking}
\end{equation}
Here, $\{h_i(\cdot)\}_{i = 1}^k$ are the set of learned input models, $\tau \in [0, 1]$ is a hyperparameter that controls the level of bias, and $\Bias(\cdot) = 1 - \Fairness(\cdot)$. For every fixed value of $\tau$, this problem achieves the the highest possible accuracy among linear combinations of learners $h \in \Hcal$; as one varies $\tau$, this problem parameterizes the \taf Curve and the fairness Pareto frontier. 

In general, solutions to \eqref{eqn:general_stacking} are not easy to directly compute, as both the $\Accuracy$ and $\Bias$ functions may be non-convex in the ensemble weights $\bw$. The use of convex surrogates for $0/1$ classification accuracy is well-established \citep{Bartlett:2006} and we take a similar approach for measures of group fairness. As \citet{Lohaus:2020} note, finding a convex surrogate for fairness is significantly more difficult than for accuracy because fairness depends on both the continuous score output by the model and the downstream decision function used to map that score onto $\{0, 1\}$ labels. Notably, this downstream decision function is often chosen independently of the stacking procedure and may be discontinuous. To sidestep these difficulties, we consider convex relaxations of \emph{score-fairness} rather than attempting to constrain the \emph{decision} fairness directly. As we will show, this transformation has sufficient fidelity to allow us to fully explore the Pareto frontier of \eqref{eqn:general_stacking} while still preserving the advantages of 
convexity.

We develop a convex notion of score-fairness for two popular group-based fairness definitions, Demographic Parity \citep{Dwork:2012,Feldman:2015} and Equality of Odds \citep{Hardt:2016}, and expect our approach to extend to other definitions of group fairness as well.  To accomplish this, we consider a unifying framework for group fairness based on ``contrast groups,'' wherein the difference in average outcomes of two groups reflects a (potential) bias to be mitigated: 
\begin{definition} \label{def:scorebias} The \emph{score bias} of a prediction system $\hat{h}: \R^p \to [0,1]$ with respect to groups $\Gcal_1, \Gcal_2$ is 
$\widetilde{\Bias} (\hat{h}) = \E[\hat{h}(\bx_i) | i \in \Gcal_1] - \E[\hat{h}(\bx_i)| i \in \Gcal_2]$
where expectations are taken with respect to the empirical measure of the data. When $\mathcal{G}_i = \{j: Z_j = i\}$, we refer to the score bias as the \emph{score deviation from demographic parity} and when $\mathcal{G}_i = \{j: Z_j = i, Y_j = 1\}$, we refer to the score bias as the \emph{score deviation from equality of opportunity} for some protected attribute $Z$ and true label $Y$, both by analogy with their decision (binary label) counterparts.
\end{definition}

Due to their linear formulation, score bias measures are particularly well-suited for use in a stacking problem. Specifically, if $\hat{H}(\cdot) = \sum w_i h_i(\cdot)$ for some fixed base learners $\{h_i\}$, then we have $\widetilde{\Bias} (\hat{H}) = \sum w_i \widetilde{\Bias} (\hat{h})$ and, more generally, $|\widetilde{\Bias} (\hat{H})| \leq \sum |w_i| |\widetilde{\Bias} (\hat{h})|$; hence, a weighted ensemble always has less score bias than its component parts, implying that model ensembles can improve fairness, in addition to their well-known improvements in accuracy. 

\begin{definition} \fairstacks, $\textsf{FS}\left(\{h_i(\cdot)\}_{i = 1}^k\right)$, is defined as the solution to the following problem:
\begin{equation}
    \argmin_{\bw \in \R^k} \sum_{j=1}^n \mathcal{L}\left(\sum_{i = 1}^k w_i h_i(\bx_j); y_j\right) 
    \text{ such that } 
    \left| \sum_{i=1}^{k} w_i \widetilde{\Bias}(h_i) \right | \leq \tau
\label{eqn:fairstacks}
\end{equation}
\end{definition}
Notice that \eqref{eqn:fairstacks} is convex for any convex loss function $\mathcal{L}$, including squared error, binomial deviance,  hinge loss, \emph{etc.} and has only linear constraints, allowing it to be easily solved at scale. Additionally, if appropriate, $\mathcal{L}$ can also include additional regularization terms not related to fairness such as a ridge penalty which may be used to reduce overfitting when the set of base learners is large. Note that in Problem \ref{eqn:fairstacks}, the range of meaningful values of $\tau$ can, and often does, extend beyond the $[0,1]$ interval. 

In light of the recent findings of \cite{Lohaus:2020} which outline some challenges with convex relaxations of fairness, it is reasonable to ask: Under what conditions does the \fairstacks problem explore the fair Pareto frontier?  While \cite{Lohaus:2020} consider pointwise differences between fairness definitions and their convex relaxations, we do not require precise pointwise bounds to fully explore the Pareto frontier.  Instead since we are trying to learn the whole \taf curve at all levels of $\tau$, all we require is that there is a monotonic relationship between score fairness and decision fairness.  This is sufficient to guarantee that our approach fully explores the Pareto frontier.  The following theorem establishes general conditions under which this monotonic relationship holds:
\begin{restatable}{theorem}{scoredecision} \label{thm:scoredecision}
Suppose the predictions of the \fairstacks problem are used to generate predicted labels via a (potentially randomized) decision function $\delta(\eta, Z)$ such that $\Delta(\eta) = \E_Z[\delta(\eta, Z)]$ is thrice-differentiable, has second derivative bounded away from zero and infinity, and is monotonically increasing in $\eta = \eta(\bx) = \sum_{i=1}^k w_i h_i(\bx)$. Then, to a second-order approximation and in expectation, the decision bias with respect to the same groups as $\widetilde{\Bias} $ is monotonically increasing in $\tau$ if
\[\frac{\bw_{\tau}^T\Cov_{i \sim \Gcal_1}[H(\bx)]\bw_{\tau}}{\bw^T_{\tau}\Cov_{i\sim\Gcal_2}[H(\bx)]\bw_{\tau}} \to \frac{\Delta''(\E_{i \sim\Gcal_2}[\bw^T_{\tau}\bx_i)])}{\Delta''(\E_{i \sim\Gcal_1}[\bw^T_{\tau}\bx_i])}\] monotonically in $\tau$, where $\bw_{\tau}$ is the solution to the \fairstacks problem at constraint level $\tau$, $H(\bx) = (h_1(\bx), h_2(\bx), \dots, h_p(\bx))$ is the vector of base model predictions at $\bx$, and $\Gcal_1, \Gcal_2$ are the groups used to evaluate both the score bias, $\widetilde{\Bias} $, and the decision bias, $\Bias$.
\end{restatable}
The conditions of Theorem \ref{thm:scoredecision} are somewhat difficult to interpret, but a simple sufficient condition is that the covariance of the base learners is equal between the two groups. This assumption is reasonable in the stacking context where the base learners are learned on the same training set and can reasonably be expected to have consistent correlation properties across both groups. Theorem \ref{thm:scoredecision} is similar to Theorem 1 of \cite{Lohaus:2020}, but gives monotonicity instead of continuity; conversely, our result holds only up to a stochastic Taylor approximation and can be weakly violated in finite samples, though we have not observed violations outside of intentionally designed counterexamples.  As an empirical example, in Figure~\ref{fig:schematics}D, score and decision fairness have a monotonic relationship. Practically, Theorem \ref{thm:scoredecision} ensures that we can explore the entire \taf/\fauc space attainable from stacking members of a given class $\Hcal$ by solving the \fairstacks problem at a fine grid of $\tau$. We hence avoid the difficulties of quantifying how accurate a particular convex relaxation is at a particular penalty $\tau$ by instead considering and making comparisons based on the entire solution set and curve. 

Next, we seek to verify whether \fairstacks achieves our stated objective of expanding the Pareto frontier of the given collection of models, $\Hcal$.  As an empirical illustration, Figure~\ref{fig:schematics}C shows that \fairstacks leads to a greatly expanded Pareto frontier compared to the frontier (\taf Curve) attained by ensemble members. In fact, the ensemble produced by \fairstacks will always exceed the \fauc of the non-stacked ensemble members, as the following proposition notes:
\begin{restatable}{proposition}{fairstackshelps} \label{prop:fairstackshelps}
Given a model class $\Hcal$, let $\textsf{FS}(\Hcal)$ be the set of models obtained by solving the \fairstacks problem \eqref{eqn:fairstacks} at all values of $\tau$. Then $\taf_{\Hcal} \leq \taf_{\textsf{FS}(\Hcal)}$ pointwise and, by extension, $\fauc^w(\Hcal) \leq \fauc^w(\textsf{FS}(\Hcal))$ for any weight function $w$ where the $\Fairness$ used to construct $\taf$ is the same score-fairness used in the \fairstacks problem. Additionally, under the conditions of Theorem \ref{thm:scoredecision}, the same inequalities hold for $\taf$ and $\fauc$ based on corresponding decision-fairness.
\end{restatable}
Putting together Proposition~\ref{prop:fairstackshelps} and Theorem~\ref{thm:convexfauc}, it is easy to see that increasing the number of the models in $\Hcal$ will always give a \fairstacks solution with larger \fauc.  This motivates us to consider ever larger collections and varieties of models as inputs to the meta-learner, \fairstacks, as we explore in the next section. Finally, though not the focus of this paper, we also present theoretical results on the out-of-sample behavior of \fairstacks in the Supplement, showing that it obtains the same attractive exponential concentration properties out-of-sample as other statistical ML 
techniques.



\begin{figure}[hpt]
  \centering
\includegraphics[width=0.48\textwidth]{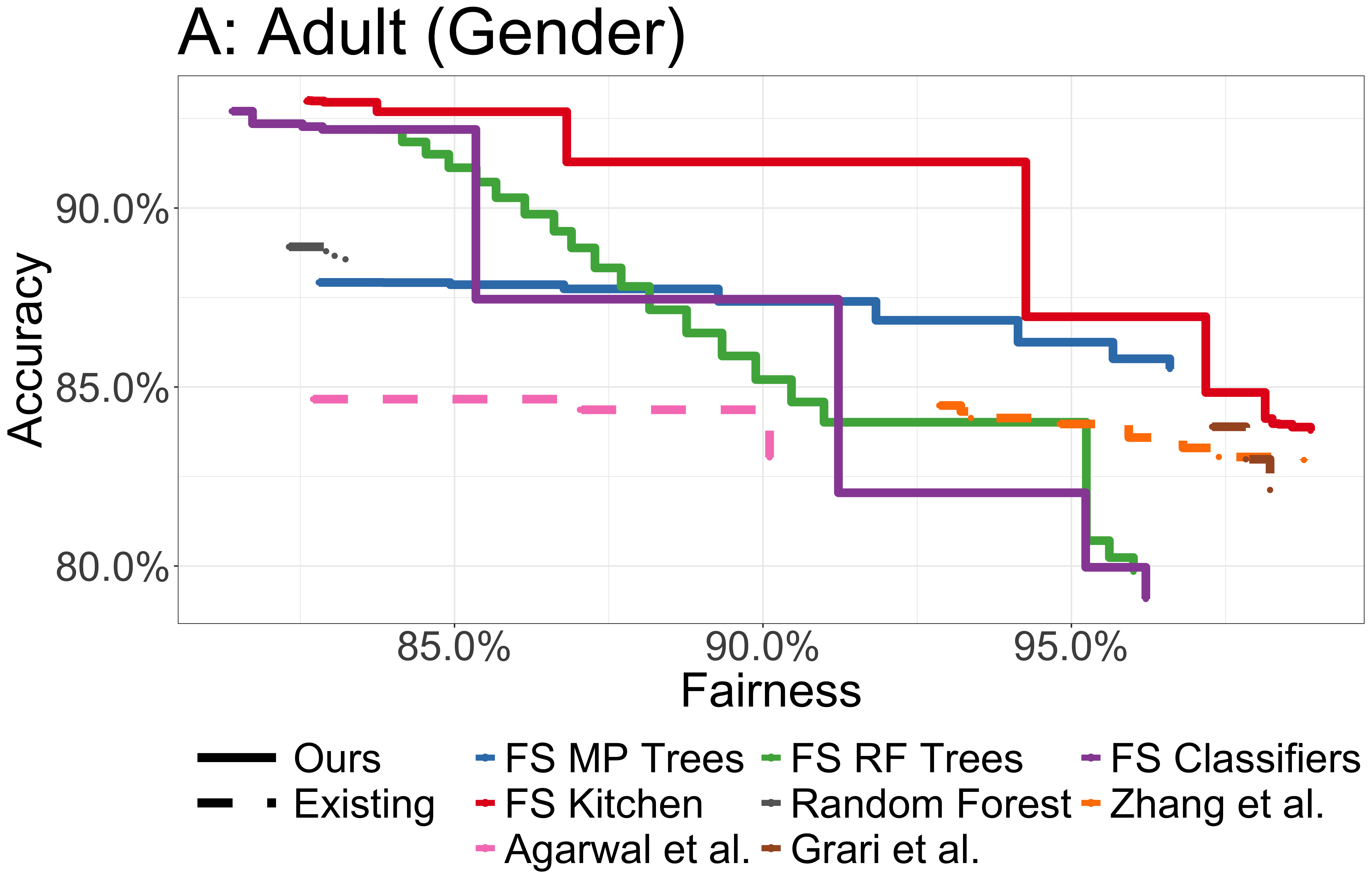} ~ \includegraphics[width=0.48\textwidth]{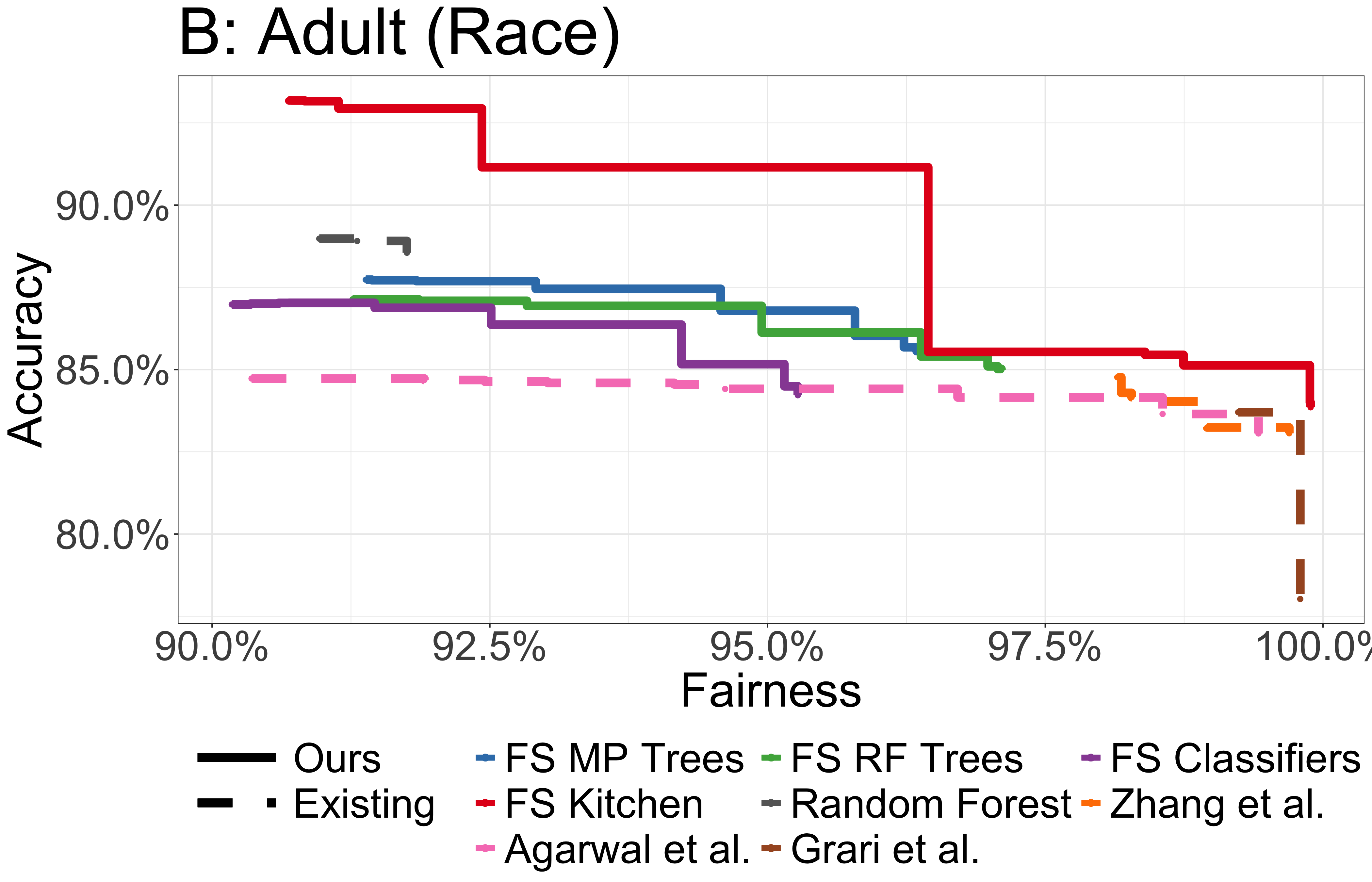} \\
\includegraphics[width=0.48\textwidth]{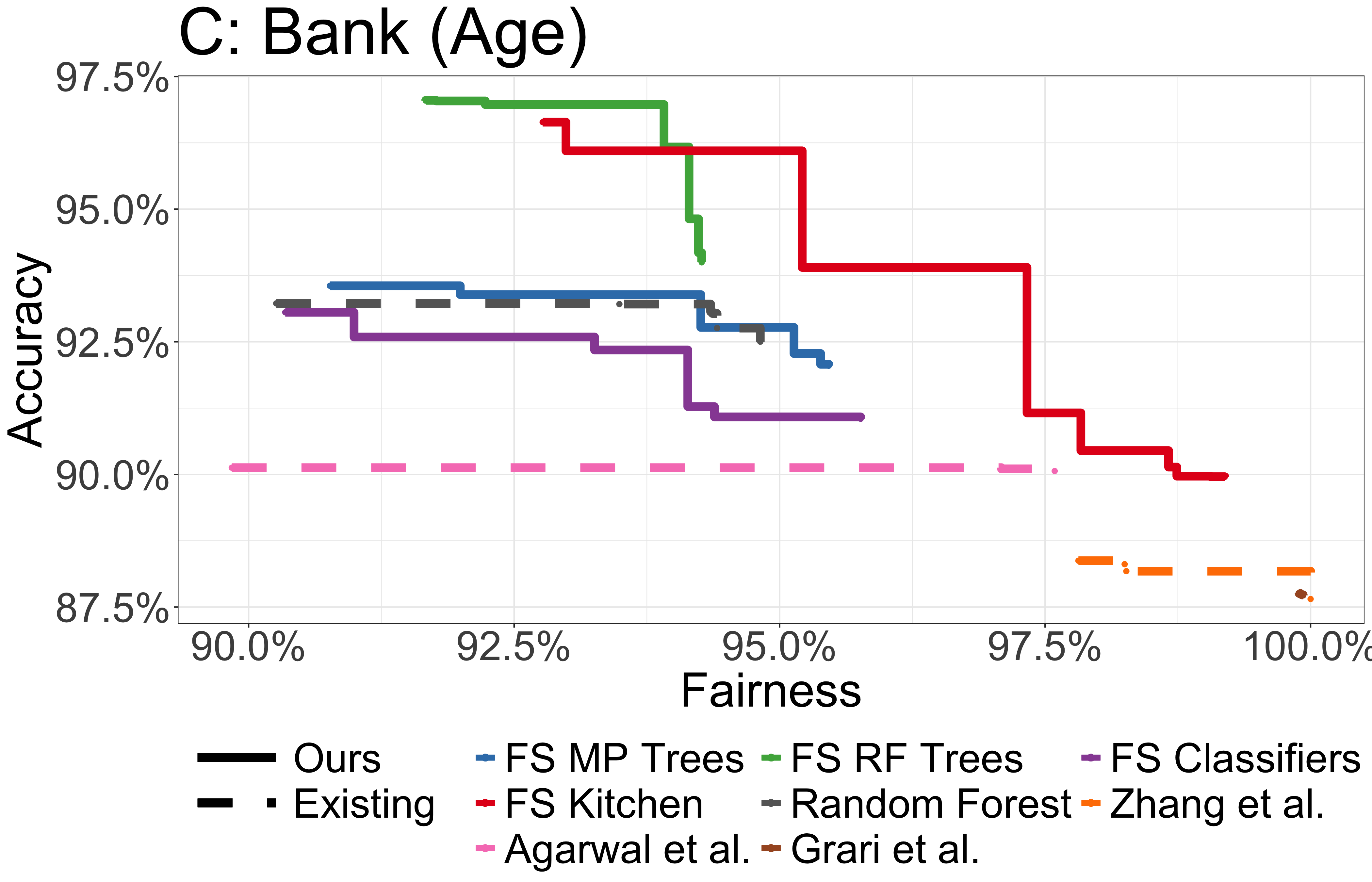} ~ \includegraphics[width=0.48\textwidth]{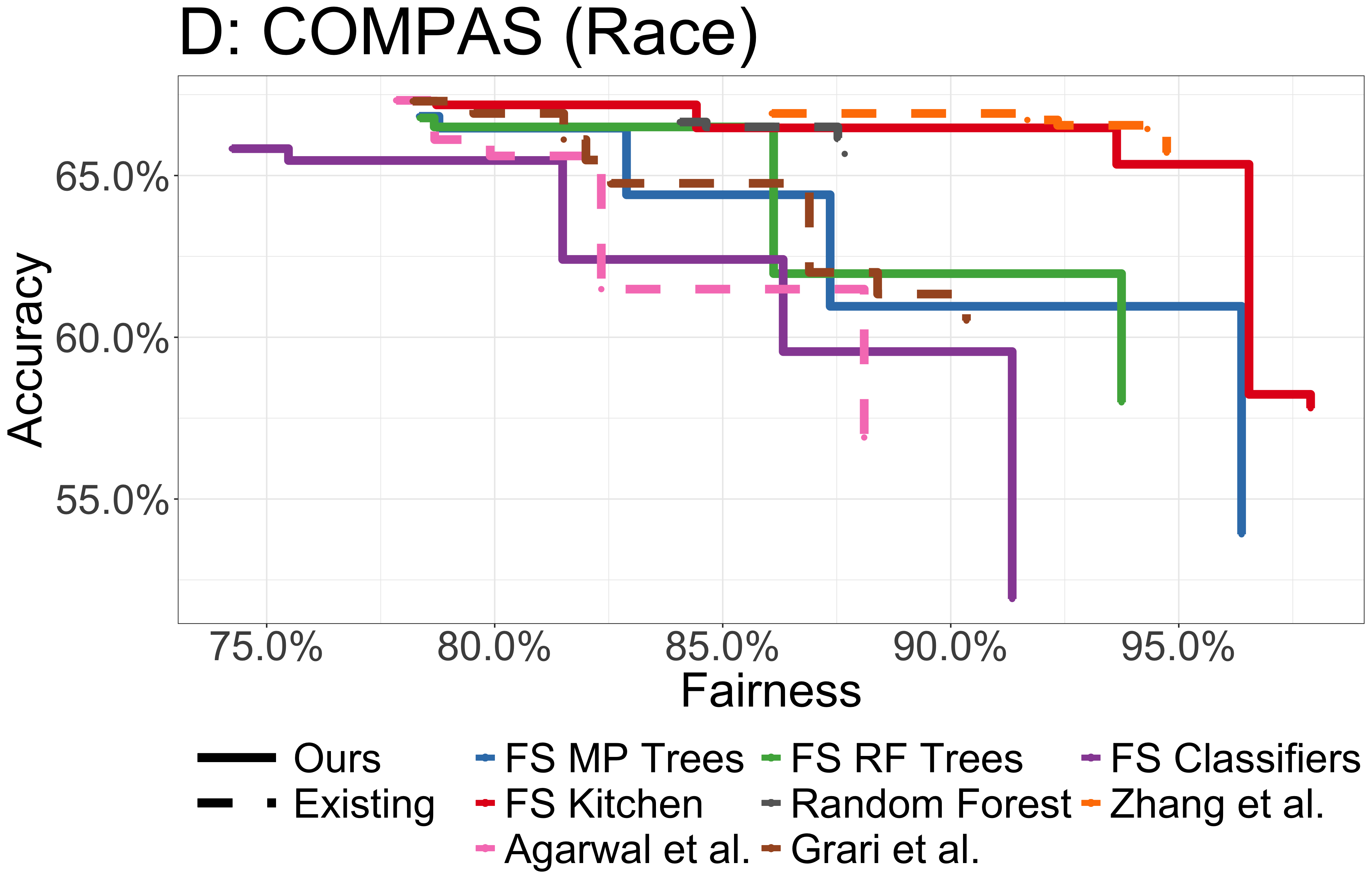} \\ 
  \caption{\taf curves from Section \ref{sec:empirical}: Demographic Parity with sensitive attribute noted in panel title. The \fairstacks \emph{kitchen-sink} (red) outperforms competing approaches, obtaining the highest \taf.}
  \label{fig:fauc}
\end{figure}

\section{Empirical Studies} \label{sec:empirical}

In this section, we demonstrate the efficacy of the \taf/\fauc framework for model comparison and the \fairstacks meta-learner for constructing fair and accurate ensembles. Our results can be found in Table \ref{tab:stackdem} and Figure \ref{fig:fauc} and clearly demonstrate the effectiveness of the proposed methodologies. We briefly describe our experiments below, with additional details and experiments in the Supplement.

\textbf{Base-Learners.} We compare several fair machine learning methodologies and ensemble families. We implement three fair classifier methods \citep{Zhang:2018,Grari:2019,Agarwal:2018} which have explicit fairness-accuracy tradeoff parameters, depicting them with dashed lines in Figure \ref{fig:fauc}. We also use the \fairstacks meta-learner to construct several weighted ensembles, shown as solid lines in Figure \ref{fig:fauc}: i) random forest trees, \emph{i.e.}, the elements of a random forest ensemble; ii) an ensemble of 1000 \emph{minipatch decision trees}, \emph{i.e.} decision trees trained on small sub-samples of features and observations \citep{Yao:2020,Toghani:2021}; iii) learners implemented in the \texttt{Scikit-Learn} and XGBoost packages \citep{Pedregosa:2011,chen:2016}; and iv) a \emph{kitchen-sink} ensemble consisting of both the stand-alone methods and the previous ensembles.

\textbf{Benchmark Datasets.} We test our methods on five standard benchmark datasets: i) the \textsf{Adult Income} dataset \citep{uci_data}, binary classification with 48,000 observations, 14 features, and 2 binary protected attributes (race and gender); ii) the \textsf{Bank} dataset \citep{uci_data}, binary classification with 45,000 observations, 16 features, and one binary protected attribute (age); iii) the \textsf{COMPAS} dataset \citep{compas_data}, binary classification with 7,000 observations, 13 features, and 2 binary protected attributes (race and gender); iv) the \textsf{Default} dataset \citep{uci_data}, binary classification with 30,000 observations, 23 features, and one binary protected attribute (gender); and the Communities and Crime (\textsf{C and C}) dataset \citep{uci_data}, a regression task with 100 features, 2,000 observations, and one binary protected attribute (race).

\textbf{Experimental Setup.} For each experiment, we split data into a 50\% training/25\% ensemble learning/25\% test split, and calculate \taf and 80\% step-weighted \fauc scores, using Demographic Parity \citep{Dwork:2012,Feldman:2015} for the fairness metric, as defined above. The threshold of $\Fairness_{\text{DP}} < 80\%$ reflects the commonly cited disparate impact threshold used by the U.S. Equal Employment Opportunity Commision (EEOC); this represents the simplest legal standard for statistical discrimination used in the U.S. and provides an impartial basis upon which to compare models.  Brier scoring \citep{Gneiting:2007}, equivalent to mean squared error, was used as the \fairstacks loss function to measure probabilistic calibration of \fairstacks predictions; a simple threshold rule at $p = 0.5$ was used for fairness assessment. \fairstacks ensembles are fit with a ridge ($\ell_2$) penalty, tuned via 5-fold cross validation within the 25\% split, to avoid overfitting. For both base learners and \fairstacks ensembles, a grid of 20 values of the tuning parameter was used. Numerical results in Table \ref{tab:stackdem} were obtained by averaging results over 10 independent data splits; quantities in parentheses are \fauc standard errors.

\textbf{Summary of Results.} Figure \ref{fig:fauc} shows \taf curves for the base learners and ensembles described above on four classification problems. While some fair learners, particularly that of \citet{Zhang:2018}, are able to obtain high accuracy, our \fairstacks framework consistently achieves a higher level of accuracy across all fairness levels. The benefits of \fairstacks are particularly pronounced for larger ensembles, with the best performance being achieved by the \emph{kitchen-sink} ensemble on all tasks. Table \ref{tab:stackdem} reports associated \fauc scores, highlighting that the high-accuracy and flexible-fairness of the \fairstacks ensembles results in the best fairness-accuracy tradeoff as measured by \fauc. Additional results in the Supplemental Materials visualize \taf curves for the data sets not shown in Figure \ref{fig:fauc}, demonstrate the use of Equality of Opportunity \citep{Hardt:2016}, and illustrate an extension of \fairstacks to multiple protected attributes, as well as providing further details of our experiments.

\begin{table}
\small
  \caption{Quantative Results for Section \ref{sec:empirical}: Demographic Parity + 80\% step-weighted \fauc. Methods in \citep{Zhang:2018,Grari:2019,Agarwal:2018} are specific to classification and cannot be applied to the \textsf{C and C} regression task.}
  \label{tab:stackdem}
\centering
  \begin{tabular}{cccccccc}
    \toprule
    \multirow{2}{*}{Method} &
      \multicolumn{2}{c}{Adult} &
      \multicolumn{1}{c}{Bank} &
      \multicolumn{2}{c}{COMPAS} &
      \multicolumn{1}{c}{Default} &
      \multicolumn{1}{c}{C and C} \\
    & Gender & Race & Age & Gender & Race & Gender & Race \\
    \midrule
    Random Forest & .772(.001) & .830(.001) & .920(.002) & .557(.004) & .524 (.003) & .848(.001)  & .623(.014) \ \\
    \citet{Zhang:2018} & .836(.020) & .845(.005) & .883(.005) & .621(.064) & .629(.018) & .778(.002) & -- \ \\
    \citet{Grari:2019} & .830(.004) & .834(.002) & .877(.013) & .590(.009)  &.641(.009)  & .850(.003) & -- \ \\
    \citet{Agarwal:2018} & .797(.003) & .841(.010) & .898(.023) & .598(.004)  & .622(.002) & .778(.014) & --\ \\
    FS MP Trees & .851(.001) & .845(.001) & .920(.002) & .621(.005) & .699(.005) & .856(.002) & .714(.004)\ \\
    FS RF Trees & .843(.001) & .851(.002) & .942(.006) &  .607(.006) & .773(.004) & .921(.002) & .695(.003)\ \\
    FS Classifiers& .823(.002) & .816(.001) & .915(.004) & .609(.004) & .735(.005) & .863(.002) & .689(.002)\ \\
    FS Kitchen Sink & \textbf{.866(.001)} & \textbf{.899(.001)} & \textbf{.947(.002)} & \textbf{.633(.003)} & \textbf{.796(.002)} & \textbf{.933(.001)} & \textbf{.734(.002)}\ \\
    \bottomrule
  \end{tabular}
\end{table}

\section{Discussion}\label{sec:discussion}

\textbf{Impact.}  
We have developed a framework for identifying, quantifying, and optimizing or expanding the empirical fairness-accuracy Pareto frontier for a collection of fitted models.  Our \taf Curves and Fairness AUC (\fauc) provide the first general, model-agnostic metric that characterize the fairness-accuracy tradeoff.  Just as ROC Curves and the associated \auc offer impartial ways to compare classifiers that balance precision and recall, our \taf Curves and \fauc offer impartial ways to compare model families that balance fairness and accuracy.  As such, our \taf Curves and \fauc provide simple, intuitive, and useful metrics that can be used to impartially compare model families for machine learning fairness problems.  This is an important contribution that fills a gap in the existing algorithmic fairness literature, providing an empirical and model-agnostic way to measure the fairness-accuracy Pareto-frontier.  Additionally, our fair model stacking framework, \fairstacks offers a simple post-processing strategy that learns an optimal weighted combination of fitted models subject to a constraint on bias.  By varying levels of the hyperparameter, this approach outlines a curve that we show expands the empirical Pareto frontier.  We also empirically show that stacking many diverse models in \fairstacks leads to major expansions of the frontier that dominate competing methods. Thus, \fairstacks can be viewed as a meta-learner that can be used to improve both the accuracy and fairness of any other learners; this is then a simple, but powerful tool in the machine learning arsenal to improve fairness. Overall, our work provides a major social impact by helping to quantify and impartially compare algorithmic fairness approaches as well as providing a useful approach to post-process and stack models that improves accuracy at every level of fairness.  

\textbf{Limitations \& Future Work.} The \taf/\fauc framework is based on existing univariate, mean-based measures of fairness and accuracy and inherits the limitations of those measures. Fundamental questions surrounding multivalent or multiple protected attributes, intersectionality, and appropriate metrics for individual fairness are active areas of research and discussion, both in machine learning and in society more broadly; as these questions are answered, the \taf framework may require extension to novel, more subtle notions of fairness. Similarly, for simple classification problems, $[0,1]$-measures of accuracy are natural, but for more complicated tasks, \emph{e.g.} those arising in ranking or computer vision, appropriate measures of accuracy are less obvious and may require alterations to our framework. In practice, fairness and accuracy must be estimated from data \citep{Lum:2022}, but we have not discussed the statistical properties of the \taf/\fauc framework that are key to rigorous model comparisons. Theorem \ref{prop:faucutility} demonstrates that a \fauc weight scheme exists for all user preferences, but does not give guidance on what weight function should be used: this decision is nuanced and likely requires consensus among a wide range of stakeholders, as well as developments in decision theory necessary to elicit and construct this consensus. A weight scheme chosen without broad input may overweight the preferences of decision makers already in power. Finally, \fairstacks uses a linear combination of models under a score-based bias constraint specific to binary groups; while we have provided sufficient conditions under which this leads to improvements in \fauc, there may be situations where tighter, but potentially non-convex, formulations perform better over a wider range of scenarios. 
As this brief discussion suggests, the \taf/\fauc framework is fertile ground for development of a fuller theory of fairness-accuracy tradeoffs.

\section*{Acknowledgements}
COL acknowledges support from the NSF Graduate Research Fellowship Program under grant number 1842494. MW's research is supported by an appointment to the Intelligence Community Postdoctoral Research Fellowship Program at the University of Florida Informatics Institute, administered by Oak Ridge Institute for Science and Education through an interagency agreement between the U.S. Department of Energy and the Office of the Director of National Intelligence. GIA acknowledges support from JP Morgan Faculty Research Awards and NSF DMS-1554821.

\printbibliography
\end{refsection}

\clearpage
\appendix 
\setcounter{figure}{0}
\renewcommand{\thefigure}{A\arabic{figure}}
\setcounter{table}{0}
\renewcommand{\thetable}{A\arabic{table}}

\begin{refsection}
\begin{center}{\LARGE \bf Supplementary Materials}\end{center}
\section{Proofs}\label{refsec:proofs}
\subsection{Proofs for Section \ref{sec:identifying}: \nameref{sec:identifying}}

The following proposition gives several properties of the \taf curve that follow immediately from the definition, but that we state here as they are useful for our subsequent discussions: 
\begin{restatable}{proposition}{tafproperties} \label{prop:tafproperties}
\taf curves are: i) closed, proper, almost-everywhere continuous, almost-everywhere differentiable, and integrable; ii) monotonically decreasing in $f$; and iii) monotonically pointwise increasing in $\Hcal$, \emph{i.e.}, if $\Hcal \subseteq \Hcal'$ then $\taf_{\Hcal}(f) \leq \taf_{\Hcal'}(f)$ for all $f \in [0, 1]$.
\end{restatable}

\begin{proof} These claims follow almost immediately from the definition of the \taf curve: 
\begin{enumerate}[label={\roman*)}]
    \item[ii)] Let $0 \leq f < f' \leq 1$. By definition, we have $\Hcal_f \subseteq \Hcal_{f'}$ where $\Hcal_f = \{h \in \Hcal : \Fairness(h) \geq f\}$. Because $\taf_{\Hcal}(f)$ is maximizing accuracy over nested sets, we have $\max_{h \in \Hcal_f} \Accuracy(f) \geq \max_{h \in \Hcal_{f'}} \Accuracy(f)$ and hence $\taf(f) \geq \taf(f')$ as desired.
    \item[iii)] Note that if $\Hcal \subseteq \Hcal'$, we have $\Hcal_f \subseteq \Hcal_f'$ for all $f$, with $\Hcal_f$ defined as above. As before,  maximization over a superset guarantees a greater maximum, so $\taf_{H}(f) \leq \taf_{H'}(f)$ for all $f$ as desired.
    \item[i)] These standard properties follow from the monotonicity of \taf and the fact that both the domain and range spaces are compact. \qedhere
\end{enumerate}
\end{proof} 

\subsection{Proofs for Section \ref{sec:quantifying}: \nameref{sec:quantifying}}
\faucutility*

Note that the topological restrictions on the preference relation $\succeq$ are weak and standard in economic theory \cite{MasColell:1995,Debreu:1983}. In essence, they require that it is possible to compare \taf curves, that the implied preferences are not inconsistent, and that the $\succeq$ relation is well behaved with respect to sequences of limits. The assumption that $\succeq$ is pointwise increasing essentially encodes that higher accuracy is preferred to lower accuracy, \emph{ceteris paribus}, at all points on the $\taf$ curve, while the assumption of preferential independence implies that the utility of increasing accuracy at fairness $f$ does not depend on accuracy at a different fairness level $f'$. The assumption that constant \taf curves have a linear utility function, \emph{i.e.} that utility of a single model is linear in its accuracy, can always be ensured by a monotonic transformation the accuracy measure used.

\begin{proof} The proof of this theorem proceeds in three parts: 
\begin{enumerate}
    \item There exists a continuous utility function $v$ encoding the preference relationship $\succeq$
    \item The utility function $v$ is linear
    \item The linear utility function $v$ can be written as a weighted sum of points on the \taf curve
\end{enumerate}

\emph{Part I: Existence of Utility.} We begin by noting that $L^2([0, 1])$ is a compact and separable topological space by standard analytic results and that the set of $\taf$ curves can be endowed with the same topology: specifically, compactness follows from compactness of $[0, 1]$ and the fact that the space of almost-everywhere continuous functions on a compact space is itself compact, while separability follows form standard results building on the denseness of $\Q$ in $\R$ and using functions defined on $\Q$ to approximate those on $\R$. From here, we invoke Theorem 1 of \citet{Mehta:1977}, noting that the both the closure and second-countability requirements of that result are satisfied by $L^2([0, 1])$ and hence by the set of $\taf$ curves which have the additional properties of almost-everywhere continuity and montonicity (\emph{cf.}, Proposition \ref{prop:tafproperties}.) \citeauthor{Mehta:1977}'s result thus gives us a utility function $v$ which is continuous with respect to $\taf$ curves. We note also that the results of \citet{Peleg:1970} could also be used here, under slightly different topological assumptions that are essentially equivalent for our purposes. 

\emph{Part II: Linearity of Utility.} Having established existence of the utility function $v$, we now argue that it can be written as a linear function of the $\taf$ curve. Our argument essentially follows that of \citet{Furken:1991} and we defer discussion of technical details to their paper.\footnote{A similar analysis also appears in the study of rational decision making in continuous-time models: see, \emph{e.g.}, the papers by \citet{Harvey:2012} and by \citet{Hara:2016}.} Essentially, we note that the existence of an additive utility function on any finite set of $\taf$ points follows from classical results of \citet{Debreu:1983}. A limiting argument extends this to \taf curves defined on the countable set $\Q$ and the denseness of $\Q$ in $\R$ and the almost-everywhere continuity of \taf curves suffices to extend to our scenario. Hence, we have 
\[v(\taf) = \int_{[0, 1]} w(f) \tilde{v}(\taf(f)) \,\text{d}f\]
for some non-negative generalized function $w$ and some pointwise utility function $\tilde{v}$, encoding the agent's utility of accuracy, independent of fairness, the existence of which is implied by the assumption of pointwise increase on $\succeq$. 

\emph{Part III: Expressing Utility in terms of \fauc.} Finally, we note that $\tilde{v}(\cdot)$ can be removed by monotonic transformation $\tilde{v}'$ such that $\tilde{v}'(\tilde{v}(x)) = x$ for all $x \in [0, 1]$. Because $\tilde{v}'$ is monotonic, it does not change the preference relation and hence \[v'(\taf) = \int_{[0, 1]} w(f) \tilde{v}'(\tilde{v}(\taf(f))) \,\text{d} f = \int_{[0, 1]} w(f) \taf(f) \,\text{d} f \propto \fauc^w\] encodes the same preference relation as $\succeq$. After linear rescaling, this gives the desired connection between $\succeq$ and $\fauc^w$. 

Note that in Part III of the proof, we used the fact that accuracy is the same quantity at all points on the \taf curve, so we could use Theorem 4 of \citet{Furken:1991} which essentially relates our problem to that of the utility of an infinite stream. The assumption about utility of constant curves ensures the existence of $\tilde{v}'$. If accuracy at various points was not comparable, we would have applied Theorem 5 of \citet{Furken:1991} and not been able to have a single (eliminable) $\tilde{v}'(\cdot)$ pointwise utility. We also note that the utility considered here is \emph{ordinal}: that is, specific values of $v$ (or of \fauc) are only useful for ordering alternatives.\qedhere
\end{proof} 

Note that the use of a generalized function $w$ in the preceding proof arises  from scenarios where the preference relation is not everywhere sensitive to $\taf$: essentially, we have to deal with the case $\langle (1, 0, \dots, 0), \bx \rangle \to x(0)$ as the grid mesh used to approximate $[0, 1]$ becomes finer. If we assume that the utility is sensitive to a non-null set of points of the $\taf$ curve, the weight term can be assumed to be a proper function.

\convexfauc*

\begin{proof} We begin by noting that the first equality $\tafi = \overline{\conv}(\taf)$ follows from basic properties of convex hulls of polytopes in the Euclidean plane. Specifically, note that
\[\conv\left[\taf \cup (0, 0) \cup (1, 0)\right]\]
is created by interpolating the vertices along the edges of the polytope. If we restrict ourselves to the upper boundary, recalling that $\overline{A} = \{(x, y): y = \sup_{(a, b) \in A: a = x} b\}$, it is clear that we have $\tafi = \overline{\conv}(\taf)$. 

For the second inequality, we first note that any point $(f, a)$ in $\tafi_{\Hcal}$ can be expressed as a linear combination of two points $(f^-, a^-)$ and $(f^+, a^+)$ corresponding to Pareto optimal elements of $\Hcal$ with the points distinct if $f \neq 0$. Let $\lambda \in [0, 1]$ be such that
\[\begin{pmatrix} f \\ a \end{pmatrix} = \lambda \begin{pmatrix} f^- \\ a^- \end{pmatrix} + (1 - \lambda) \begin{pmatrix} f^+ \\ a^+ \end{pmatrix}\]
The existence of this two-point representation is guaranteed by Carath\'eodory's Theorem on convex bodies, restricting attention to the 1-dimensional face of $\overline{\conv}(\taf)$ containing $(f, a)$. Now note that if $h^-, h^+ \in \Hcal$ are the models with fairness and accuracy $(f^{\pm}, a^{\pm}$) respectively, the randomized combination $h = \lambda h^- + (1 - \lambda) h^+$ has the desired fairness and accuracy: specifically, note that
\begin{align*}
    \Accuracy(h) &= \E[\Accuracy(h)] \\
    &= \E[Z\Accuracy(h^-) + (1 - Z) \Accuracy(h^+)] \\
    &= \E[Z]\Accuracy(h^-) + \E[1-Z]\Accuracy(h^+) \\
    &= \lambda f^- + (1 - \lambda) f^+ \\
    &= f
\end{align*}
as desired for any $\Accuracy$ measure which is an average over observations. The same argument holds for $\Fairness$, showing that every point in $\tafi$ is obtained by a convex combination of elements of $\Hcal$.
\end{proof}

\subsection{Proofs for Section \ref{sec:optimizing}: \nameref{sec:optimizing}}
As discussed in the main text, by constraining the scores, \emph{e.g.}, the linear predictor $\eta = \bX\beta$ term of logistic regression, rather than the decisions of the \fairstacks problem, we are able to retain convexity and computational tractability. On its own, however, score fairness is not enough to imply decision fairness: consider two populations where scores are distributed as a point mass at 0.6 for one group and a equal mixture of two point masses at 0.4 and 0.8 for the other group ($\eta | A = 0 \sim \delta_{0.6}$ vs $\eta | A = 1 \sim \frac{1}{2}\delta_{0.4} + \frac{1}{2}\delta_{0.8}$). In both cases, $\E[\eta] = 0.6$, but if $\eta$ is used as input to a decision function which thresholds at $0.5$, we see that $P[\hat{Y} = 1 | A = 0] = 1$ while $P[\hat{Y} | A = 1] = 0.5$, resulting in a significant violation of demographic parity. 

While trivial, this example highlights what is essentially the only failure mode of score fairness: mean scores matching while higher moments (\emph{e.g.}, variance) not matching between classes. We note however that this failure cannot occur in the \fairstacks context: for a fixed data generating process, shrinking the means of the two classes together shrinks all higher moments as well or, at least, does  not pull them apart, resulting in increased decision fairness as the \fairstacks score constraint is tightened.  
\scoredecision*

\begin{proof} We begin by taking a second-order Taylor-expansion of $\Delta$ around $\eta_0 $ to find: 
\[\Delta(\eta) = \Delta(\eta_0) + \Delta'(\eta_0) (\eta - \eta_0) + \frac{1}{2} \Delta''(\eta_0) (\eta - \eta_0)^2 + \mathcal{O}(\Delta'''(\eta_0))\]
Substituting $\eta_0 = \bw_{\tau} ^TH(\bx) $ and taking expectations over $H(\bx) $ on both sides ($\eta_0 = \bw_{\tau} ^T\E[H(\bx) ]$) we have: 
\begin{align*}
    \E[\Delta(\bw_{\tau} ^TH(\bx) )] &= \Delta(\E[\bw_{\tau} ^TH(\bx) ]) + \frac{1}{2}\Delta''(\E[\bw_{\tau} ^TH(\bx) ]) \Cov(\bw_{\tau} ^TH(\bx) ) + \mathcal{O}(\Delta'''(\bw^T_{\tau}\E[H(\bx) ])))\\ &=  \Delta(\E[\bw_{\tau} ^TH(\bx) ]) + \frac{1}{2}\Delta''(\E[\bw_{\tau} ^TH(\bx) ])\bw_{\tau} ^T\Cov(H(\bx) )\bw_{\tau}  + \mathcal{O}(\Delta'''(\bw^T_{\tau}\E[H(\bx) ]))
\end{align*}
The difference in this quantity between two sub-groups ($\Gcal_1, \Gcal_2$) gives the expected decision bias induced by decision rule $\Delta$ for a \fairstacks solution of $\bw_{\tau}$: 
{\small \begin{align*}
\text{Bias}(\bw_{\tau}; \Gcal_1, \Gcal_2) &= \overbrace{\Delta(\E_{H(\bx)  \sim \Gcal_1}[\bw_{\tau} ^TH(\bx) ]) - \Delta(\E_{H(\bx)  \sim \Gcal_2}[\bw_{\tau} ^TH(\bx) ])}^{A}\\ &\quad+ \frac{1}{2}\underbrace{\bw_{\tau} ^T\left[\Delta''(\E_{H(\bx)\sim \Gcal_1}[\bw_{\tau} ^TH(\bx) ])\Cov_{H(\bx)\sim\Gcal_1}(H(\bx) ) - \Delta''(\E_{H(\bx)\sim \Gcal_2}[\bw_{\tau} ^TH(\bx) ])\Cov_{H(\bx)\sim\Gcal_2}(H(\bx) )\right]\bw_{\tau} }_B \\ &+ \mathcal{O}(\Delta'''(\bw^T_{\tau}\E[H(\bx) ])))
\end{align*}}

Clearly the zeroth-order term, $A$, is decreasing in $\tau$ by monotonicity of $\Delta$ and the \fairstacks constraint, which forces the two group means towards each other. Controlling the second order term, $B$, requires slightly stronger assumptions: rearranging 
\[\bw_{\tau} ^T\left[\Delta''(\E_{H(\bx)\sim \Gcal_1}[\bw_{\tau} ^TH(\bx) ])\Cov_{H(\bx)\sim\Gcal_1}(H(\bx) ) - \Delta''(\E_{H(\bx)\sim \Gcal_2}[\bw_{\tau} ^TH(\bx) ])\Cov_{H(\bx)\sim\Gcal_2}(H(\bx) )\right]\bw_{\tau} \to 0\]
gives the convergence condition in the statement of the theorem, as desired. \qedhere
\end{proof} 

Three special cases of the preceding analysis are worth noting: 
\begin{enumerate}[label={\Roman*.}]
    \item If the covariances of the base learners are the same across groups, then $B$ reduces to
    \[B = \left[\Delta''(\E_{\bx\sim\Gcal_1}[\bw^T_{\tau}H(\bx)]) - \Delta''(\E_{\bx\sim\Gcal_2}[\bw^T_{\tau}H(\bx)])\right]\bw^T_{\tau} \Cov(H(\bx))\bw_{\tau}\]
    which clearly goes to $0$ as $\tau \to 0$ by the same argument as the zeroth order term. This is the simple sufficient condition stated in the main text.
    \item A weaker sufficient condition for decision bias to vanish, up to second order, as $\tau \to 0$ is for $\bw_{\tau}$ to converge to the nullspace of 
    \[\frac{\Delta''(\E_{\bx \sim \Gcal_1}[\bw^T_{\tau}H(\bx)])}{\Delta''(\E_{\bx \sim \Gcal_2}[\bw^T_{\tau}H(\bx)])}\Cov_{\bx\sim\Gcal_1}(H(\bx)) - \Cov_{\bx\sim\Gcal_2}(H(\bx))\]
    That is, we do not actually need the covariances to match asymptotically: we only need $\bw_{\tau}$ to lie in the nullspace of their difference. In essence, this only requires that the stacked ensemble is ``unaware'' of the difference in variances, not that  no difference exists.
    \item If we can bound $\|\Cov_{\bx\sim \Gcal_1}(H(\bx)) - \Cov_{\bx\sim \Gcal_2}(H(\bx))\|_{\text{op}}$ above by $C$, then we can bound $B$ above by 
    \[C|\Delta''(\E_{\bx\sim\Gcal_1}[\bw^T_{\tau}H(\bx)]| + |\Delta''(\E_{\bx\sim\Gcal_1}[\bw^T_{\tau}H(\bx)]-\Delta''(\E_{\bx\sim\Gcal_2}[\bw^T_{\tau}H(\bx)]|\|\Cov_{\bx\sim\Gcal_2}(H(\bx))\|_{\text{op}}\]
    The second term goes to $0$ as $\tau \to 0$, but the first term is essentially constant in $\tau$. This highlights the behavior described in the text before this proof: if there is a systemic difference in the covariance structure across groups, we cannot guarantee that decision bias goes to zero. We do note, however, that it is essentially monotonic\footnote{Specifically, when we move from $B$ to an upper bound on $B$, we loose traditional monotonicity and have the following weaker form of monotonicity: $\text{Bias}(\bw_{\tau}; \Gcal_1, \Gcal_2)$ is upper bounded by some sequence $\kappa_{\tau}$ such that $\kappa_{\tau}$ is monotonically decreasing as $\tau \to 0$.} and the qualitative results of the discussion in Section \ref{sec:optimizing} still hold.
\end{enumerate}

Theorem \ref{thm:scoredecision} is similar to Theorem 1 of \citet{Lohaus:2020}, but with three major differences: i) we establish (approximate) monotonicity of a specific constraint as opposed to continuity with respect to a broader class of constraints; ii) we do not assume strong convexity of the problem; and iii) we allow an arbitrary decision function $\delta$ to be used, rather than a threshold at $0$. Conversely, our result holds only up to a higher-order Taylor approximation and in expectation and can be weakly violated in finite samples, though we have not observed violations outside of intentionally designed counterexamples. 

\fairstackshelps*

\begin{proof} Let $\textsf{Stack}(\Hcal)$ be all linear combinations of models in $\Hcal$. Clearly $\Hcal \subsetneq \textsf{Stack}(\Hcal)$ so, by Proposition \ref{prop:tafproperties}(iii), we have $\taf_{\Hcal} \leq \taf_{\textsf{Stack}(\Hcal)}$. To extend this result to $\textsf{FS}(\Hcal)$ instead of $\textsf{Stack}(\Hcal)$, we note that for a fixed fairness level $\tau$, 
$\textsf{Stack}(\Hcal)_{\tau} = \textsf{FS}(\Hcal)_{\tau}$ where $\Hcal_f = \{h \in \Hcal: \Fairness(h) \geq f\}$, because taking models with bias at most $1 - \tau$ is equivalent to taking models with fairness at least $\tau$. Since this inequality holds for all $\tau$, we have $\taf_{\textsf{Stack}(\Hcal)} = \taf_{\textsf{FS}(\Hcal)}$ and hence $\taf_{\Hcal} \leq \taf_{\textsf{FS}(\Hcal)}$ as desired. 

The conditions of Theorem \ref{thm:scoredecision} then let us extend this result to suitable decision-fairness measures because pointwise dominance is preserved under monotonic transformation of the shared abscissa ($x$-axis). We note that the conditions of Theorem \ref{thm:scoredecision}, or conditions substantively equivalent, are necessary to ensure the fairness measures used in solving \fairstacks and in constructing the resulting \taf curves are congruent.
\end{proof}

\subsection{Out-of-Sample Performance of \fairstacks}
\begin{theorem} Let $\bw_{\tau}$ be the solution to the \fairstacks problem, trained on $n_1, n_2$ samples from groups $\Gcal_1, \Gcal_2$ respectively, and let $t(\bx) : \R^p \to \{0, 1\} = \bone_{\bw_{\tau}^T H(\bx) > c}$ be a thresholding (decision) rule for some fixed $c$. If $t(\bx)$ has classification accuracy $A_{\text{train}}$ and bias $B_{\text{train}}$ on the training data used to create the ensemble weights, then
\begin{align*}
    A_* &\leq A_{\text{train}} + 2\sqrt{\frac{(p + 1)\log(2 \min\{n_1, n_2\} + 1)}{\min\{n_1, n_2\}}} + \sqrt{\frac{\log 1/\delta}{\min\{n_1, n_2\}}} \\ 
    B_* &\leq B_{\text{train}} + 2\sqrt{\frac{(p + 1)\log(2\min\{n_1, n_2\} + 1)}{\min\{n_1, n_2\}}} + \sqrt{\frac{\log 1/\delta}{\min\{n_1, n_2\}}}
\end{align*}
each with probability at least $1 - \delta$, where $A_*, B_*$ are the expected out-of-sample classification accuracy and bias respectively. 

Furthermore, if the same classifier, $t(\cdot)$, is run on a test sample of $n_1', n_2'$ samples from groups $\Gcal_1, \Gcal_2$ respectively, then 
\begin{align*}
    A_{\text{test}} &\leq A_{\text{train}} + \sqrt{\frac{\log 1/\epsilon}{4\min\{n_1', n_2'\}}} + 2\sqrt{\frac{(p + 1)\log(2 \min\{n_1, n_2\} + 1)}{\min\{n_1, n_2\}}} + \sqrt{\frac{\log 1/\delta}{\min\{n_1, n_2\}}} \\
    B_{\text{test}} &\leq B_{\text{train}} + \sqrt{\frac{\log 1/\epsilon}{2\min\{n_1', n_2'\}}} + 2\sqrt{\frac{(p + 1)\log(2 \min\{n_1, n_2\} + 1)}{\min\{n_1, n_2\}}} + \sqrt{\frac{\log 1/\delta}{\min\{n_1, n_2\}}}
\end{align*}
each with probability at least $1  - \delta - \epsilon$, where $A_{\text{test}}, B_{\text{test}}$ are the realized test classification accuracy and bias respectively.
\end{theorem}

\begin{proof} For simplicity, we assume that both groups have $n = \min\{n_1, n_2\}$ samples in the training data and $n'=\min\{n_1', n_2'\}$ samples in the test data. Marginally tighter results can be obtained by treating $n_1, n_2$ separately, but at the cost of more cumbersome analysis that yields little additional insight. Having additional samples will only tighten the concentration results used, so the results claimed are true in the general case. With this simplification, we reduce the problem to the analysis of a classifier trained on $2n$ samples and can use the well-developed tools of empirical risk minimization and VC dimension; specifically, we rely on the tools presented by \citet{Boucheron:2005}. 

Combining their Equation (3), Theorem 3.4, and the remark at the end of Section 3, we have the general result that, with probability at least $1 - \delta$, 
\[\sup_{f \in \Fcal} |P_{2n}f - Pf| \leq 2\sqrt{\frac{(p + 1)\log(2n + 1)}{n}} + \sqrt{\frac{\log 1/\delta}{n}}\]
where the supremum is taken over all linear threshold functions\footnote{We have an extra VC dimension because we allow for an arbitrary threshold (bias/intercept term) rather than fixing the threshold at 0.}, $P_{2n}$ refers to the empirical expectation of $f$ over the training data and $P$ refers to the expected value under \textsc{iid} sampling. Substituting this bound into Equation (2) of \citet{Boucheron:2005},  we obtain the first set of results in our theorem: note that nothing in their analysis requires that $L$ actually be the loss function used, though this is commonly the case in the analysis of classification accuracy, so we can use the bias function here as well without issue. 

We now turn to bounds for accuracy and bias on a finite test set: in this case, we rely on standard Hoeffding-type bounds for concentration of Bernoulli random variables. Specifically, if we have $2n'$ samples each of which are correctly classified with probability $A_*$, then the empirical classification accuracy $A_{\text{test}}$  
is simply the mean of $2n$ Bernoulli random variables each with probability $A^*$. Hoeffding's inequality, which can be found as Theorem 2.2.6 in the book by \citet{Vershynin:2018}, then implies that
\[\mathbb{P}\left(A_{\text{test}} - A^* \geq t\right) \leq e^{-4t^2n'}\]
Combining this with the previous result and setting $\epsilon = e^{-4t^2n}$, we find that
\[A_{\text{test}} \leq A_{\text{train}} + \sqrt{\frac{\log 1 / \epsilon}{4n'}} + 2\sqrt{\frac{(p + 1)\log(2n + 1)}{n}} + \sqrt{\frac{\log 1/\delta}{n}}\]
with probability at most $1 - \delta - \epsilon$ as desired. 

For the bias term, we have essentially the same analysis, but with $n$ (arbitrarily paired) observations each of which have different labels (a ``bias occurence'') with probability $B^*$. Repeating the argument from above, Hoeffding's inequality gives, 
\[\mathbb{P}\left(B_{\text{test}} - B^* \geq t\right) \leq e^{-2t^2n'}\]
Combining this with the previous result and setting $\epsilon = e^{-2t^2n'}$, we find that 
\[B_{\text{test}} \leq B_{\text{train}} + \sqrt{\frac{\log 1 / \epsilon}{2n'}} + 2\sqrt{\frac{(p + 1)\log(2n + 1)}{n}} + \sqrt{\frac{\log 1/\delta}{n}}\]
as desired. \qedhere
\end{proof}

Note that the $\log(2\min\{n_1, n_2\}+1)$ terms can be removed at the cost of a worse leading constant: see Theorem 3.4 of \citet{Boucheron:2005} for details. Also note that for certain fairness measures, the ``effective sample size'' in this result ($\min\{n_1, n_2\}$) may be a significant underestimate: specifically, if Equality of Opportunity \citep{Hardt:2016} or other fairness measures that depend only on a subset of the samples are used, the accuracy (depending on the full sample) may concentrate more quickly than these results would suggest.

\section{Additional Material for Section \ref{sec:empirical} - \nameref{sec:empirical}}\label{refsec:empirical}

\subsection{Additional Experimental Details} \label{app:details}
\subsubsection{Data Sets and Preprocessing} \label{app:data}
In this section, we explain mores specific details about the data sets used in our experiments. 
\begin{itemize}
    \item \textbf{UCI Adult Income} \citep{uci_data, Kohavi:1996} Target values state whether a person's income is $\geq \$50k $ given a series of attributes. Originally, the data set contains 9 categorical features and 6 continuous features with $n=48,222$ observations ($p=99$ total features after one-hot encoding). It has two binary protected attributes: Gender (male or female) and Race (white or non-white). We discarded the \texttt{fnlwgt} variable, which indicates the number of people census takers believe that observation represents.
    
    Note that we used the ``classic'' version of the Adult Income data set based on the 1994 U.S. census: \citet{Ding:2021} note weakness of this data set and give extensions and updates based on more recent censuses.
    
    \item \textbf{Bank}\citep{uci_data, Moro:2014} Target values state whether a client has subscribed to a term deposit. The data set contains $p=16$ features ($p = 58$ after one-hot encoding) on $n=44,388$ clients of a Portuguese banking institution. The protected attribute is age, encoded as a binary attribute indicating whether the client's age is between 33 and 63 years or not. Continuous variables were left as is and categorical variables were converted into one-hot encoded vectors.
    
    \item \textbf{COMAS Recidivism}\citep{compas_data} Target values state whether or not the individual recidivated within 2 years of their most recent crime. This dataset contains $p=13$ binary features and  $n=5,278$ observations. It has two binary protected attributes: Gender (male or female) and Race (white or non-white).
    
    We note that this data set is controversial on its own merits and in its application: we refer the reader to \citet{Bao:2021} and to references therein for a thorough discussion of the limitations of this data set.
    
    \item \textbf{Default}\citep{uci_data,Yeh:2009}  Target values that state  whether an individual will default on payments. This data set contains $p=23$ features and $n=30,000$ observations. Gender is the sensitive attribute, encoded as a binary attribute, male or female. Continuous variables were left as is and categorical variables were converted into one-hot encoded vectors.
    \item \textbf{Communities and Crime}\cite{uci_data} Target values state the normalized \emph{per capita} violent crime rates in various communities in the United States. It contains $p=104$ features and $n=1,969$ observations. Race is the protected attribute encoded as a binary attribute, majority white community or not. We removed features that had $80 \%$ or greater missing values.

\end{itemize}

\subsubsection{Hyper-Parameter Selection and Model Tuning} \label{app:tuning}

While we analyze \fairstacks in its constrained form (\emph{cf.}, Equation \eqref{eqn:fairstacks}), we use the following equivalent penalized form in our numerical experiments: 

\begin{align*}
    \argmin_{\bw \in \R^k} \sum_{j=1}^n \mathcal{L}\left(\sum_{i = 1}^k w_i h_i(\bx_j); y_j\right) + \lambda^2\left( \sum_{i=1}^{k} w_i \widetilde{\Bias}(h_i) \right)^2
\end{align*}
Note that we square the score bias term ($\widetilde{\Bias}$) to yield a differentiable penalty term, making the penalized form particularly easy to solve using a damped Newton method. 

Here $\lambda \in \R_{\geq 0}$ is a hyper-parameter controlling the degree of bias: higher values of $\lambda$ give less biased (more fair) solutions, with $\lambda = \infty \Leftrightarrow \tau = 0$ giving a perfectly fair ensemble. We use a grid of 20 log-spaced values of $\lambda \in [10^0, 10^6]$ to construct \taf curves for our \fairstacks ensembles. To avoid overfitting our ensemble weights, we also include a ridge-regression type penalty ($+\frac{\alpha}{2}\|\bw\|_2^2$) with $\alpha \in [10^2, 10^7]$ chosen to optimize the five-fold cross validated estimate of the \fauc score. (Note that cross-validation was performed within the model stacking step: reported test \fauc scores are indeed unbiased.) 

In our comparisons with the methods of \citet{Zhang:2018, Agarwal:2018,Grari:2019}, we use standard publicly available implementations:
\begin{itemize}
\item \textbf{Adversarial Debiasing - \citet{Zhang:2018}}: Adversarial debiasing attemps to learn a classifier under an objective function which balances prediction accuracy and an adversary's ability to determine the protected attribute(s) from the predictions. We use the implementation from the \texttt{AI Fairness 360} \citep{aif360} package\footnote{\url{https://github.com/Trusted-AI/AIF360}} and vary the fairness-accuracy tradeoff parameter (the weight given to the adversary in the composite objective) over a linearly-spaced grid on the $[0, 1]$ interval. Default settings are used for all other hyper-parameters. 

\item \textbf{Reduction-Based Fair Classification - \citet{Agarwal:2018}}: The reductions approach to fair classification reweights or relabels that data to find the most accurate and fair version of the input classifier. We use the implementation from the \texttt{Fairlearn} package\footnote{\url{https://fairlearn.org/v0.5.0/api_reference/fairlearn.reductions.html}} and vary the fairness-accuracy tradeoff parameter (the weight given to the adversary in the composite objective) over a linearly-spaced grid on the $[0, 1]$ interval. Default settings are used for all other hyper-parameters.

\item \textbf{Fair Adversarial Gradient Tree Boosting - \citet{Grari:2019}}: The Fair Adversarial Gradient Tree Boosting methodology attempts to learn an ensemble of tree-based classifiers using a variant of gradient boosting which balances prediction accuracy with an adversary's ability to determine the protected attribute(s) from the predictions. We use the authors' reference implementation\footnote{\url{https://github.com/vincent-grari/FAGTB}} and vary the fairness-accuracy tradeoff parameter (the weight given to the adversary in the boosting objective) over a linearly-spaced grid on the $[0, 0.2]$ interval. Default settings are used for all other hyper-parameters. 
\end{itemize}

\subsubsection{Computational Resources Used} \label{app:computational}
Computational experiments were run on a shared server with a AMD Ryzen Threadripper 3970X 32-Core Processor (64 Logical CPUs / 32 Physical CPUs) at 3700 MHz and 252 GB of memory. No GPU resources were used. Moderate parallelization (20 threads) was used to perform our simulation studies, taking approximately 2 days of total wall clock time ($<1000$ hours of CPU time) for all results in this paper.

Note that, because \taf identification is quasi-linear and the \fairstacks meta-learner is convex, computational costs are dominated by training base learners and not by the proposed methods of the paper. For the results shown here, Algorithm \ref{alg:one} and \fairstacks both took negligible time ($<10$ seconds) in all our experiments, even for our largest ensembles, with time dominated by the calculations necessary to estimate base learner fairness and accuracy.

\subsection{Additional Experimental Results}
\subsubsection{Demographic Parity - Additional Data Sets}
\begin{figure}[H]
  \centering
  \includegraphics[width=0.48\textwidth]{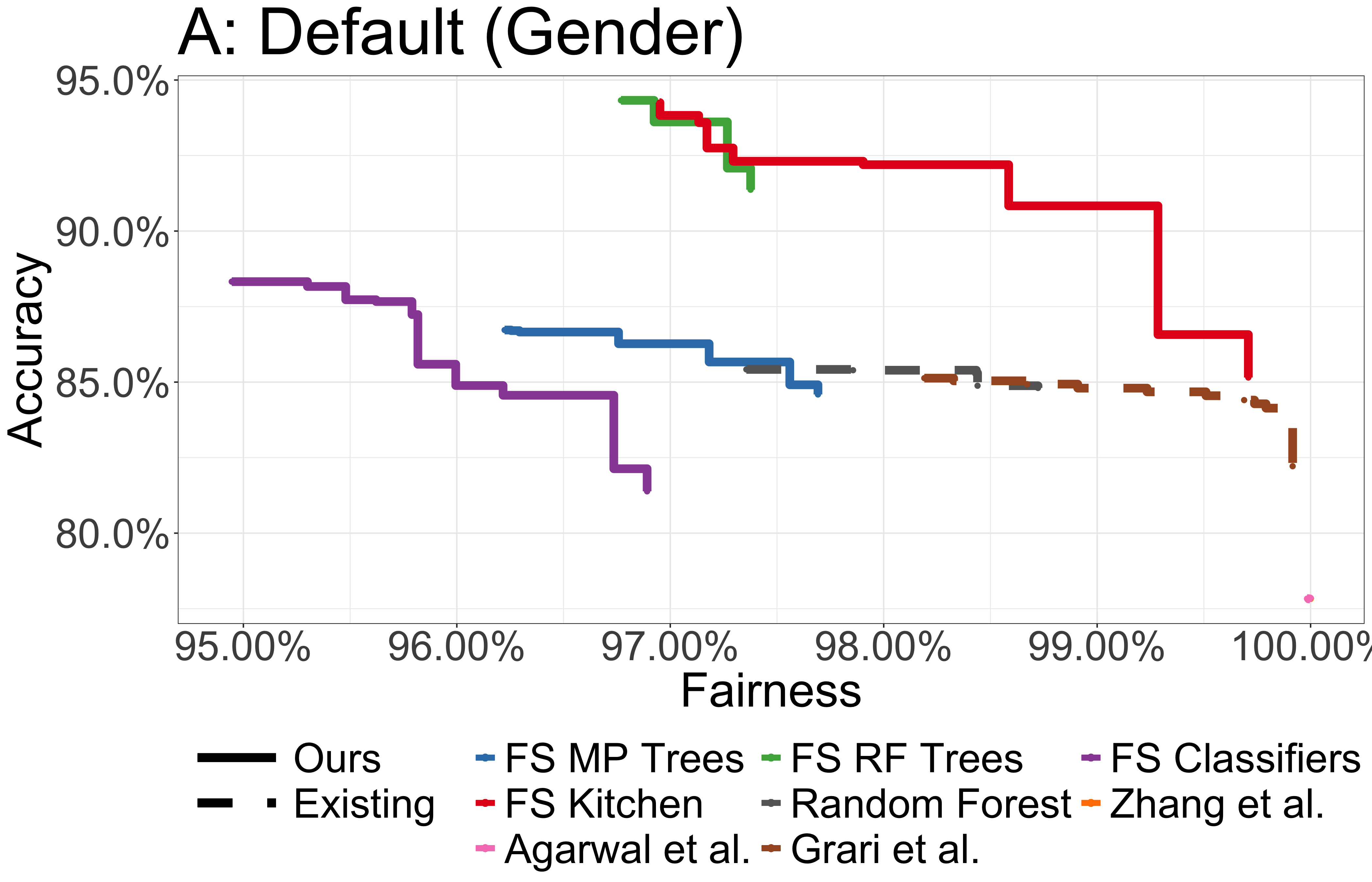}
  \includegraphics[width=0.48\textwidth]{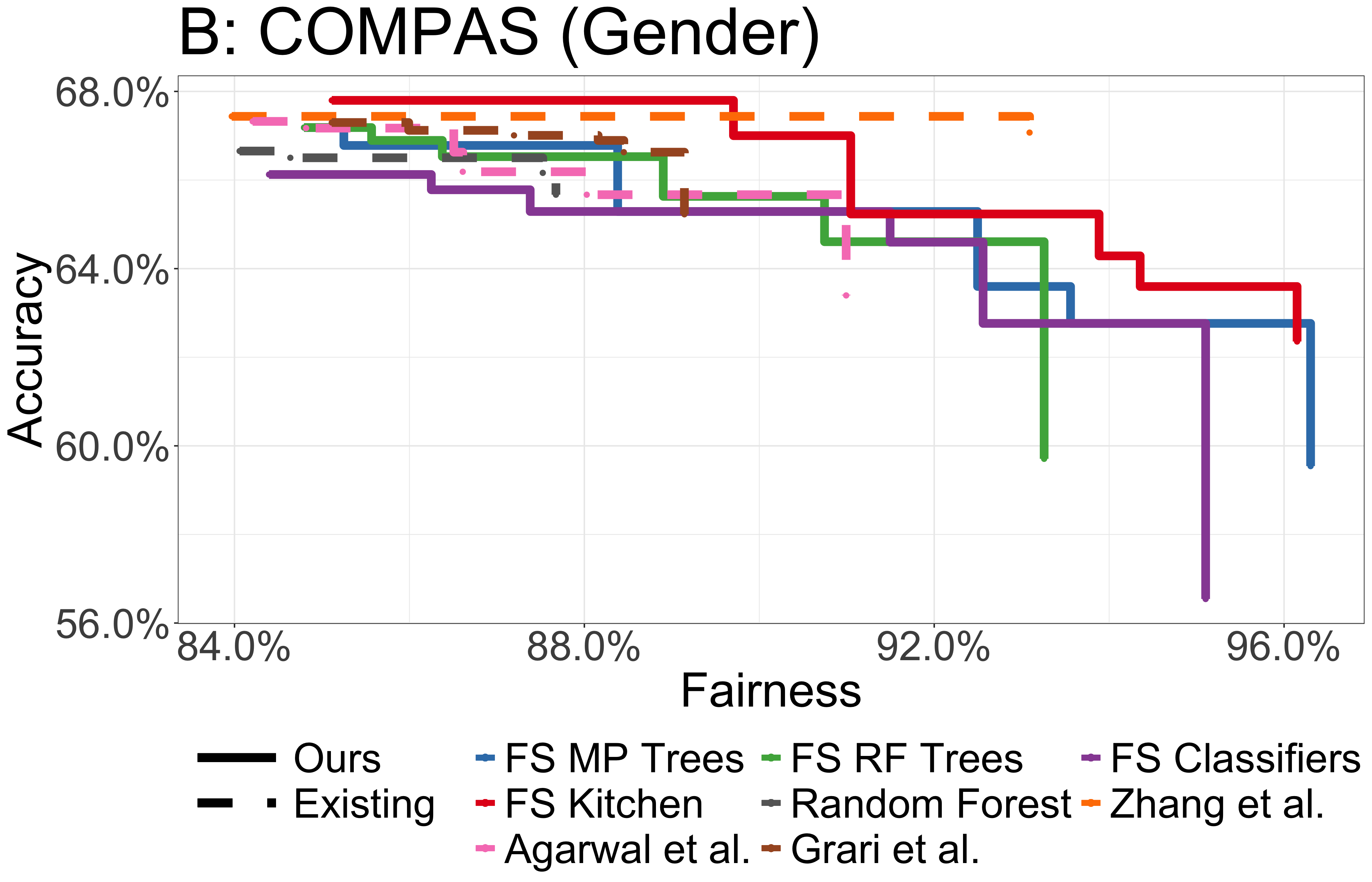}
  \includegraphics[width=0.48\textwidth]{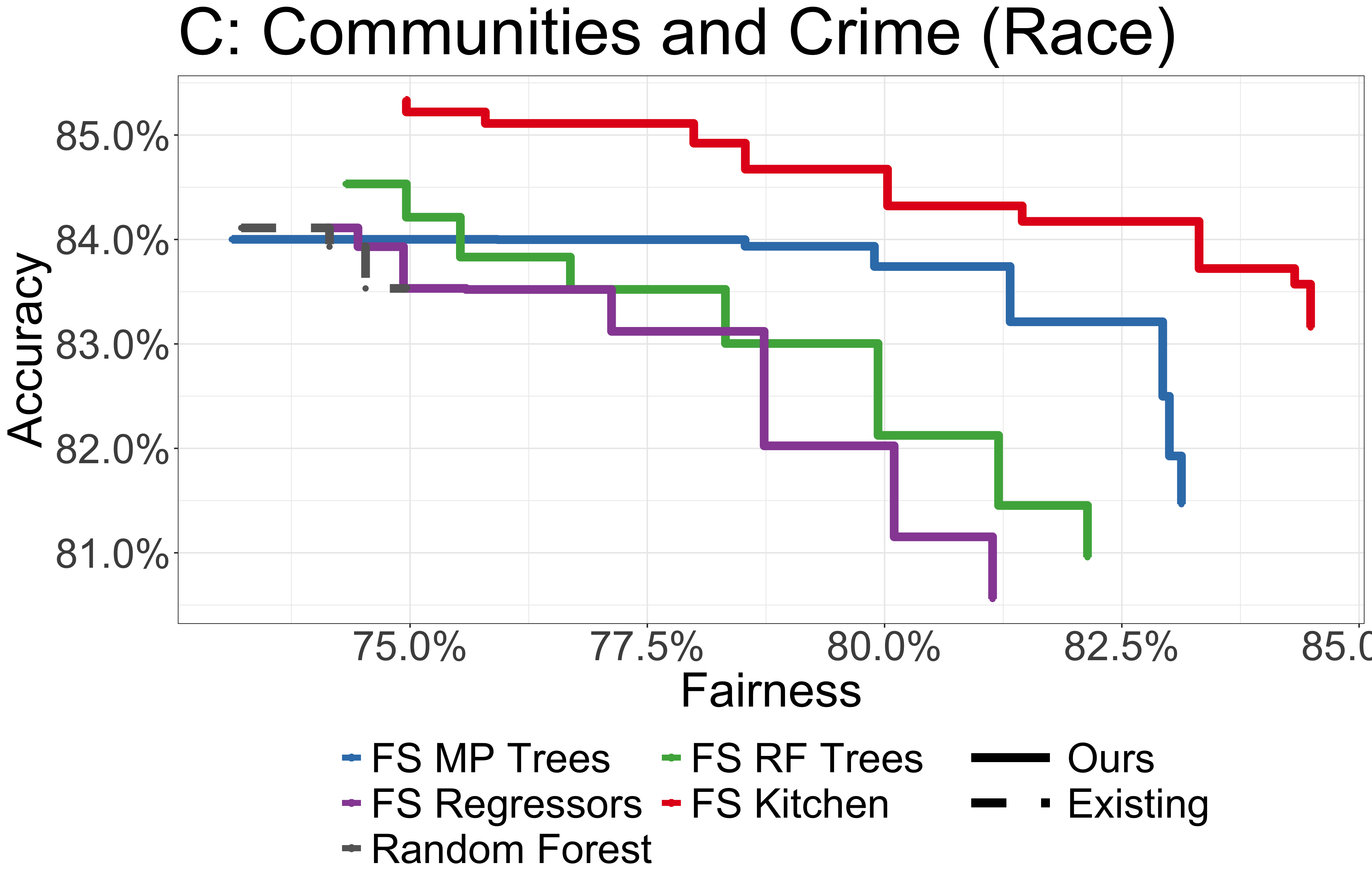}
  
  \caption{\taf curves for three fair ML tasks not shown in Figure \ref{fig:fauc}: see discussion in Section \ref{sec:empirical} and Sections \ref{app:data}-\ref{app:tuning} for more details of our experimental approach. We see that on all tasks, the largest \fairstacks ensemble (\emph{kitchen-sink}) achieves the highest accuracy at all fairness levels, yielding the highest overall \fauc scores (\emph{cf.}, Table \ref{tab:stackdem}). Note that the methods of \citet{Zhang:2018,Grari:2019,Agarwal:2018} are omitted for the regression (\textsf{C and C}) task as those methods do not natively support regression.}
  \label{fig:taf_additional}
\end{figure}

Figure \ref{fig:taf_additional} visualizes \taf curves for data sets not appearing in Figure \ref{fig:fauc} of the main text of the paper.  Note that panel C (the \textsf{C and C} data set) has fewer curves because the methods of \citet{Zhang:2018,Grari:2019,Agarwal:2018} cannot be applied to this regression task. See Figure \ref{fig:fauc} in the main text for other data sets considered.

\subsubsection{\fauc Table: Equality of Opportunity} \label{app:fauceo}
Our results in Section \ref{sec:empirical} are presented for Demographic Parity, but the \taf/\fauc framework and the \fairstacks meta-learner can be used for other fairness measures. In this section, we repeat our analysis from above, now using Equality of Opportunity (EO) \citep{Hardt:2016} as our fairness metric. Equality of Opportunity is defined by the criterion $\E[\hat{Y} | A = 1, Y = 1] = \E[\hat{Y} | A = 0, Y = 1]$ and we measure fairness as deviation from this ideal: $\Fairness_{\text{EO}} = 1 - |\E[\hat{Y} | A = 1, Y = 1] - \E[\hat{Y} | A = 0, Y = 1] |$, where $A$ is the protected attribute, $Y$ is the ground truth label, and $\hat{Y}$ is the predicted label of a classifier. For our score bias measure in \fairstacks, we use the difference in subgroup means, where the same subgroups are used as calculating fairness. (\emph{cf}., Definition \ref{def:scorebias}).

Table \ref{tab:stackeo} reports the EO-\fauc scores of several base learners and ensemble construction strategies. As before, we use a weight function at 80\% to provide an unbiased comparison, while enforcing conformance with prevailing U.S. legal standards (\emph{cf.} Table \ref{tab:stackdem}). As with our Demographic Parity-based experiments, we see that the \fairstacks framework consistently achieves the highest \fauc scores across all data sets considered: additionally, consistent with our theoretical results, \fairstacks fit on the largest set of base learners (\emph{kitchen-sink}) obtains the best Pareto frontier.  


\begin{table}[H]
  \caption{Quantative Results for Section \ref{app:fauceo}: Equality of Opportunity + 80\% step-weighted \fauc. We omit the \textsf{C and C} regression task because it is unclear how to define Equality of Opportunity for continuous labels.}
  \label{tab:stackeo}
  \centering
  \begin{tabular}{ccccccc}
    \toprule
    \multirow{2}{*}{Method} &
      \multicolumn{2}{c}{Adult} &
      \multicolumn{1}{c}{Bank} &
      \multicolumn{2}{c}{COMPAS} &
      \multicolumn{1}{c}{Default} \\
    & Gender & Race & Age & Gender & Race & Gender  \\
    \midrule
    Random Forest & .782(.001) & .814(.001) & .900(.002) & .568(.003) & .557(.003) & .834(.010) \ \\
    \citet{Zhang:2018} & .825(.030) & .840(.007) & .872(.004) & .618(.052) & .617(.061) & .845(.003) \ \\
    \citet{Grari:2019} & .826(.003) & .821(.004) & .889(.012) & .627(.005)  & .628(.004)  & .789(.002) \ \\
    \citet{Agarwal:2018} & .800(.004) & .802(.002) & .883(.011) & .621(.003) & .629(.003) & .791(.010)\ \\
    FS MP Trees & .861(.001) & .857(.001) & .914(.004) & .639(.004) & .683(.002) & .891(.002) \ \\
    FS RF Trees & .852(.001) & .849(.003) & .918(.015) &  .622(.004) & .700(.005) & .921(.003) \ \\
    FS Classifiers& .839(.002) & .837(.001) & .907(.003) & .601(.005) & .712(.006) & .883(.002) \ \\
    FS Kitchen Sink & \textbf{.871(.001)} & \textbf{.863(.001)} & \textbf{.929(.003)} & \textbf{.641(.004)} & \textbf{.738(.003)} & \textbf{ .944(.001)} \ \\
    \bottomrule
  \end{tabular}
\end{table}

\subsubsection{\fauc Table: Multiple Protected Attributes}
Our \fairstacks framework can be used to improve fairness for multiple protected attributes simultaneously by adding a constraint for each protected attribute. Though this sort of intersectional fairness \citep{Buolamwini:2018} is not the primary focus of our paper, we give a brief demonstration here.  

Table \ref{tab:both} reports \fauc scores for both gender- and race-fairness for four different \fairstacks ensembles generated from Random Forest trees: an unconstrained ensemble, \fairstacks  with a gender-fairness constraint, \fairstacks with a race-fairness constraint, and \fairstacks with both constraints. That is, we solve the \fairstacks problem with various constraints and compute \fauc scores for each protected attribute separately. As in our other experiments, we use a 80\%-step function weight scheme in calculating \fauc. 

As can be seen in Table \ref{tab:both}, \fairstacks always yields better \fauc scores for both protected attributes. As expected, \fairstacks achieves a better \fauc score for the group used in the constraint, but doubly-constrained \fairstacks is highly competitive in all scenarios. This suggests that, at least for the data sets considered here, \fairstacks can achieve fairness for multiple groups without significant decreases in accuracy. While this result is promising, we leave development and analysis of a ``composite'' \fauc, quantifying the tradeoff between both fairness measures and accuracy, to future work. 

\begin{table}[hbt]
  \caption{Application of \fairstacks to Multiple Protected Attributes: \fauc scores in columns two and four are calculated  using Gender as the protected attribute, while scores in columns three and five use Race as the protected attribute. While the presence of any fairness constraint improves \fauc scores for both protected attributes, the multiply-constrained \fairstacks achieves near-optimal performance in all scenarios considered.}
 \label{tab:both}
 \centering
  \begin{tabular}{ccccc}
    \toprule
    \multirow{2}{*}{Constraint} &
      \multicolumn{2}{c}{Adult} &
      \multicolumn{2}{c}{COMPAS} \\
    & Gender & Race & Gender & Race  \\
    \midrule
    None & .772(.001) & .830(.001) & .557(.004) & .524(.003) \ \\
    Gender & .843(.001) & .829(.002) &  .607(.006) & .641(.004)   \ \\
    Race & .817(.002) & .851(.002) & .592(.005) & .773(.004)   \ \\
    Both & .849(.001) & .841(.002) & .600(.005)& .727(.004)\ \\
    \bottomrule
  \end{tabular}
\end{table}

\printbibliography[title={Additional References}]
\end{refsection}
\end{document}